\documentclass{seeleai-report}

\usepackage{multirow}
\usepackage{colortbl}
\usepackage{rotating}

\definecolor{c_header}{HTML}{5A2E85}
\definecolor{c_causal}{HTML}{F6F4FA}
\definecolor{c_bidir}{HTML}{E7F6DC}
\definecolor{c_mask}{HTML}{FAFAFC}
\definecolor{eva_purple}{HTML}{5A2E85}
\definecolor{eva_lavender}{HTML}{F6F4FA}
\definecolor{eva_green}{HTML}{9DFF3F}
\definecolor{eva_mint}{HTML}{E7F6DC}

\title{EVA01: Unified Native 3D Understanding and Generation via Mixture-of-Transformers}
\author{SeeleAI Team}
\date{\today}
\seeleShortTitle{EVA01 Technical Report}
\seeleIconLink{\seeleLinkLogo}{https://www.seeles.ai/}
\seeleIconLink{\seeleLinkPageIcon}{https://www.seeles.ai/research/pages/EVA01}

\begin{document}
\maketitle

\begin{abstract}
This paper addresses the challenge of integrating 3D mesh as a native modality within Multimodal Large Language Models (MLLMs).
Diffusion-based large reconstruction models decouple semantic understanding from geometric reasoning, operating as stateless reconstructors conditioned on dense 2D pixel priors.
Recent MLLM-based methods treat the 3D modality as an external output rather than a native component of the multimodal sequence, making incremental adaptations without systematic analysis of how geometric manifolds align with MLLM feature spaces.
We introduce EVA01, a unified framework that extends the modality boundary of MLLMs to natively incorporate 3D mesh understanding, generation, and context-aware editing.
Built upon a Mixture-of-Transformers (MoT) architecture, EVA01 decouples the model into a pre-trained Understanding Expert ($E_{\text{und}}$) and a structurally mirrored Generation Expert ($E_{\text{gen}}$), coupled through shared global self-attention with hard modality routing.
This design aligns the semantic latent space of the MLLM backbone with the geometric manifold, enabling direct transfer of multimodal priors without intermediate 2D representations.
Results show that EVA01 achieves state-of-the-art native text-to-3D generation fidelity and unlocks robust long-context multi-turn geometric editing with identity preservation---a capability fundamentally inaccessible to stateless reconstruction pipelines.
Our findings further offer architectural insights for integrating 2D foundation models with 3D tasks, informing the design of 3D-native multimodal systems.
\end{abstract}

\begin{center}
  \includegraphics[width=\linewidth]{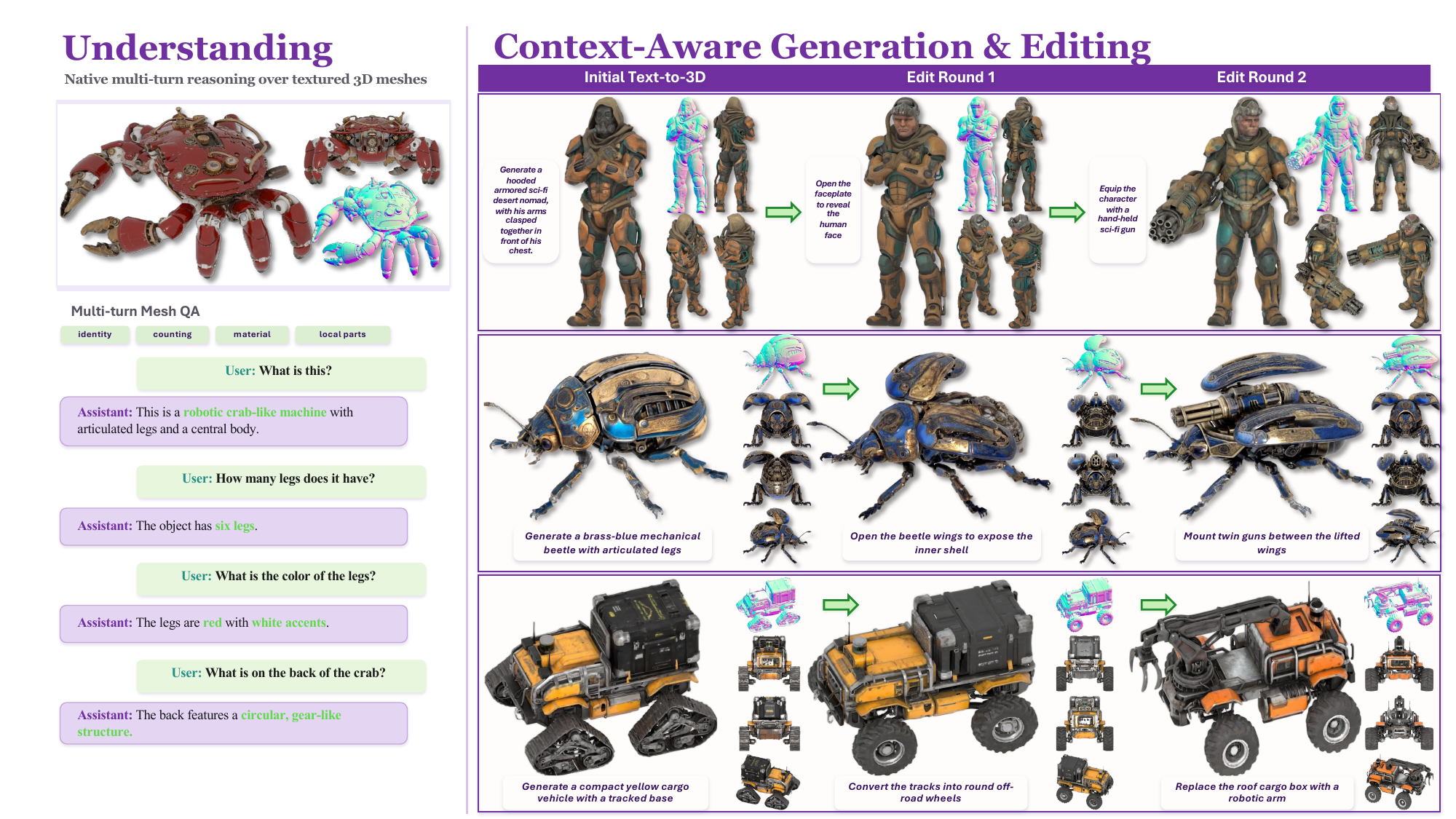}%
  \captionsetup{hypcap=false}
  \captionof{figure}{
  \textbf{Native Multi-Turn 3D Understanding, Generation, and Editing.}
  EVA01 unifies mesh-native 3D understanding, text-conditioned generation, and context-aware editing within a continuous context.
  Left: multi-turn question answering over a textured mesh.
  Right: text-to-3D generation followed by sequential edits across three assets.
  Each trajectory applies localized structural edits---revealing a face, adding weapons, unfolding wings, replacing tracks with wheels, attaching a robotic arm---while preserving object identity across turns.
  All edits are generated without explicit masks, conditioned on the full interaction history.
  }
  \label{fig:teaser}
\end{center}

\section{Introduction}

Diffusion-based large reconstruction models have driven recent progress in 3D content creation by leveraging dense pixel-level features from multi-view images to reconstruct 3D geometry.
These methods---spanning Score Distillation Sampling (SDS)~\cite{poole2023dreamfusion,wang2023prolificdreamer}, cascaded multi-view reconstruction~\cite{xu2024instantmesh,hunyuan3d22025tencent}, and feed-forward large reconstruction models~\cite{xiang2024structured}---exploit the spatial priors encoded in 2D foundation models~\cite{rombach2022ldm,saharia2022imagen,peebles2023dit} to achieve high visual fidelity.
However, they decouple semantic understanding from geometric reasoning: semantic interpretation is delegated to a frozen image encoder, while geometric construction is treated as a downstream lifting operation.
This renders them geometry reconstructors conditioned on dense pixel priors rather than generative models that reason over 3D structure.
Consequently, they operate in a stateless manner---every edit requires full re-generation, with no mechanism to preserve geometric identity across sequential modifications, limiting their utility for iterative 3D design.

Recent works have explored incorporating Multimodal Large Language Models (MLLMs) into 3D generation to bridge this semantic-geometric gap~\cite{ye2025shapellm,ye2026omni123,huang2026cg,chen2025sar3d,huang2026unimesh,chen2026know3d}.
While these approaches advance language-driven 3D generation, they treat the MLLM primarily as a semantic feature extractor or conditioning module without systematic representation-level analysis of the relationship between MLLM feature spaces and the 3D geometric manifold.
Existing methods have not yet demonstrated a post-training pipeline---with modality alignment, progressive curriculum learning, and expert decoupling---that follows the scaling paradigm established by unified image understanding and generation models~\cite{deng2025bagel,xie2024showo,wu2024janus}.
This limits their capacity to achieve robust semantic-geometric alignment and to scale with model capacity and data volume.

Native image generation models such as Nano-Banana and GPT-Image have recently demonstrated that treating images as first-class tokens within a unified sequence unlocks consistent, multi-turn understanding and editing inaccessible to prior approaches.
Can the same transition be realized for 3D?
This requires integrating 3D mesh as a first-class modality within the MLLM sequence stream, which introduces a representational challenge beyond simple multimodal feature concatenation.
A unified sequence stream must accommodate three modalities with structurally distinct properties: text encodes abstract semantics with no spatial inductive bias; images capture dense pixel-level spatial correlations that implicitly encode projective geometry~\cite{oquab2023dinov2,darcet2023vitneedreg,simeoni2025dinov3}; and 3D meshes impose strict topological constraints---manifoldness, genus, and local connectivity---that must be respected at every sequence position.
The core question is therefore not merely how to generate 3D shapes from language, but how to align these three heterogeneous modalities within a unified sequence representation---where text and images follow autoregressive modeling while mesh geometry is generated via flow matching---such that semantic intent, visual grounding, and geometric validity are jointly preserved.
This alignment problem is especially acute under the scarcity of large-scale, high-quality text--3D paired data~\cite{deitke2022objaverse,deitke2023objaverse_xl,zhang2025texverse,chang2015shapenet}, demanding training strategies that efficiently transfer multimodal priors to the 3D domain.

We introduce \textbf{EVA01}, a context-aware unified MLLM that natively integrates 3D mesh understanding, generation, and multi-turn editing within a single Mixture-of-Transformers (MoT) architecture~\cite{liang2025mixtureoftransformers}.
EVA01 decouples the model into two complementary experts: a pre-trained \emph{Understanding Expert} ($E_{\text{und}}$) that serves as a stable semantic anchor preserving the multimodal priors of the MLLM backbone, and a structurally mirrored \emph{Generation Expert} ($E_{\text{gen}}$) dedicated to geometry synthesis.
Through shared global self-attention with hard modality routing, $E_{\text{gen}}$ explicitly queries semantic representations from $E_{\text{und}}$, enabling cross-modal knowledge transfer while maintaining optimization independence.
To address long-context 3D editing---where strict topological constraints must be preserved across sequential modifications---we construct a large-scale interleaved text--image--mesh dataset and formulate 3D generation as a \textbf{multimodal sequence modeling} task, where each geometric state is predicted conditioned on both the current instruction and the full historical context.
This stateful formulation enables \textbf{context-aware, identity-preserving} multi-turn 3D editing inaccessible to stateless reconstruction pipelines.

Our contributions are threefold:
\begin{itemize}
\item \textbf{Unified MoT-based 3D MLLM:}
  To our knowledge, the first native 3D MLLM to combine mesh understanding, generation, and context-aware multi-turn editing
  within a single Mixture-of-Transformers architecture, integrating mesh as a first-class modality via a
  structurally mirrored generation expert and shared global self-attention for cross-modal
  knowledge transfer.

\item \textbf{Curriculum-Based Semantic Alignment:}
  A multi-stage post-training strategy that bridges the mismatch between semantic
  representations and geometric structures via progressive modality alignment, using interleaved
  text--image--mesh sequences and modality dropout to establish robust cross-modal correspondences.

\item \textbf{Stateful Editing Paradigm:}
  A stateful generation formulation that models 3D editing as conditional sequence modeling,
  achieving identity-preserving geometric modifications across multi-turn interactions and providing
  insights into the training dynamics of 3D-native MLLMs.
\end{itemize}

\begin{figure}[t]
  \centering
  \includegraphics[width=\linewidth]{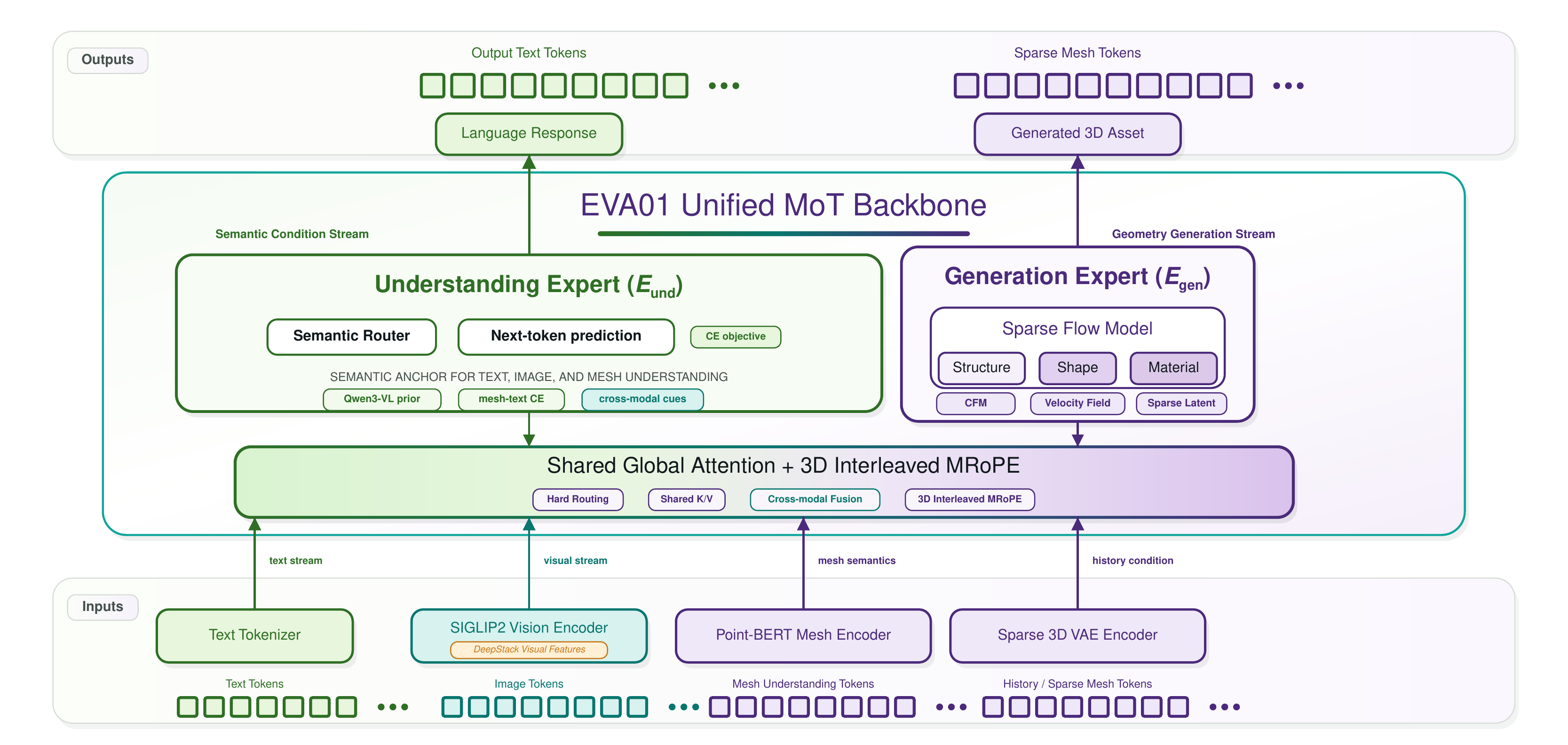}
  \caption{
  \textbf{The Architecture of EVA01.}
  EVA01 organizes tokenized text, image, and mesh inputs within a unified \textbf{Mixture-of-Transformers} backbone.
  The \textcolor{eva_green}{\textbf{Understanding Expert}} ($E_{\text{und}}$) preserves the semantic priors of the pre-trained MLLM through Qwen tokenization, SigLIP2 visual encoding, Point-BERT mesh encoding, and DeepStack visual features.
  The \textcolor{eva_purple}{\textbf{Generation Expert}} ($E_{\text{gen}}$) consumes history-conditioned \textbf{Sparse Mesh Tokens} and predicts the structured 3D latent through \textbf{Structure}, \textbf{Shape}, and \textbf{Material} stages under Conditional Flow Matching.
  A shared fusion card, \textbf{Shared Global Attention + 3D Interleaved MRoPE}, enables cross-modal routing from semantic conditions to geometry generation while preserving spatial correspondence for context-aware editing.
}
  \label{fig:architecture}
\end{figure}

\section{Related Work}

\subsection{3D Generative Models and Latent Representations}
Generating high-fidelity 3D geometry at scale remains challenging due to data sparsity, cubic volumetric complexity, and strict topological requirements~\cite{kazhdan2006poisson,lorensen1987marchingcubes}.
Optimization-based distillation (e.g., SDS) and cascaded multi-view reconstruction leverage 2D diffusion priors for 3D synthesis~\cite{poole2023dreamfusion,wang2023prolificdreamer,xu2024instantmesh}, but treat 3D as a reconstruction task conditioned on dense pixel priors rather than native generation, often producing view inconsistencies and decoupled semantic reasoning~\cite{yang2024hunyuan3d,hunyuan3d22025tencent}.
Native 3D diffusion methods improve scalability by operating within learned latent manifolds, compressing structured signals via 3D VAEs for transformer-based generation~\cite{xiang2024structured,hunyuan3d22025tencent,wu2025unilat3d,jia2025ultrashape}.
However, existing latent representations face a trade-off: global token sets oversmooth high-frequency details~\cite{zhang2023shapetovecset}, while sparse localized tokens incur substantial computational cost~\cite{lai2025latticedemocratizehighfidelity3d}.
Disjoint geometry--appearance modeling further causes semantic misalignment and fragile surfaces during remeshing, motivating unified, structurally-aware representations for coherent end-to-end generation~\cite{jia2025ultrashape,wu2025unilat3d,lai2025latticedemocratizehighfidelity3d}.

\subsection{Unified Multimodal Large Language Models}
Multimodal Large Language Models (MLLMs) have evolved from LLMs augmented with modality-specific encoders to unified architectures that integrate generation within the language backbone, enabling fine-grained control and compositional reasoning across long contexts~\cite{deng2025bagel,dong2024dreamllm,xie2024showo,wu2024janus,ma2024janusflow,chen2025januspro}.
Extending this paradigm to 3D, however, is constrained by geometric topology and the need for spatial consistency.
Unlike 2D media, 3D assets require strict identity preservation across sequential modifications; cascaded text-to-image-to-3D pipelines decouple semantic planning from geometric construction, leading to identity drift and topological discontinuities.
Effective unification further demands latent representations expressive enough to capture high-frequency detail yet compatible with autoregressive prediction.
Building a practical 3D-native multimodal model therefore requires synergizing MLLM semantic priors with scalable, topologically valid latent representations and production-ready data pipelines, enabling direct, context-aware, identity-preserving editing aligned with linguistic intent~\cite{wu2025unilat3d,jia2025ultrashape,lai2025latticedemocratizehighfidelity3d,han2024mvimgnet2}.

\subsection{Unified 3D Multimodal Large Models}

ShapeLLM-Omni~\cite{ye2025shapellm}, built on Qwen2.5-VL-Instruct-7B, treats 3D as a first-class modality by expanding the LLM vocabulary with 8,192 learned 3D VQ-VAE tokens~\cite{xiang2024structured} and performing understanding and generation via fully autoregressive next-token prediction. Its single unified backbone processes text, image, and 3D tokens without modality-specific expert decoupling, causing pre-trained MLLM priors to degrade under the conflicting optimization demands of semantic reasoning and geometric synthesis; moreover, discrete VQ-VAE tokenization inherently limits geometric fidelity.
Omni123~\cite{ye2026omni123} unifies text-to-2D and text-to-3D generation via dual-stream attention, Cube3D VQ-VAE tokenization, and dual text encoders (CLIP~+~Qwen3); it supports instruction-based editing and achieves strong results on Edit3D-Bench, yet each edit proceeds as an independent forward pass without persistent geometric identity across turns.
CG-MLLM~\cite{huang2026cg}, built concurrently on Qwen3-VL, shares a similar MoT backbone with dedicated understanding and generation experts under hard modality routing, making it the closest architectural parallel to EVA01. The critical differences lie in representation and data: it relies on a VecSet-based 3D representation~\cite{hunyuan3d2025hunyuan3d}---which our ablations (Sec.~\ref{subsec:ablation}) show is substantially weaker than a structured sparse grid---performs mesh understanding on rendered 2D views rather than native mesh tokens, and lacks a multi-stage post-training curriculum with modality-specific optimization schedules.
Sar3d~\cite{chen2025sar3d} proposes next-scale autoregressive prediction over a multi-scale 3D VQ-VAE for fast generation (0.82\,s per object), and repurposes truncated token scales for captioning via a separate LLaMA; generation and understanding thus rely on distinct model components rather than a natively unified MLLM.
UniMesh~\cite{huang2026unimesh} employs Bagel~\cite{deng2025bagel} as a frozen MLLM conditioner---outputting FLUX VAE image latents bridged through a LoRA-fine-tuned Mesh Head to Hunyuan3D's implicit shape decoder---and introduces Chain-of-Mesh for inference-time iterative editing. Know3D~\cite{chen2026know3d} follows a three-stage pipeline: Qwen2.5-VL for semantic reasoning, Qwen-Image-Edit-2511 (20B MMDiT) for back-view generation, and frozen TRELLIS.2 with parallel cross-attention injection of intermediate MMDiT hidden states~\cite{xiang2024structured}. In both cases the MLLM serves as an external conditioner rather than a native reasoning backbone processing text, images, and 3D mesh tokens within a shared sequence space.

In summary, while these methods advance the integration of MLLMs with 3D content, none combine three capabilities that define EVA01: (1) a MoT architecture that decouples semantic understanding from geometric generation with shared cross-modal attention, (2) native 3D mesh tokens processed within the MLLM sequence stream via a structured grid-based latent representation and flow matching, and (3) stateful, long-context multi-turn 3D editing where each geometric update is conditioned on the full interaction history with explicit identity preservation.

\section{Methodology}

EVA01 is a unified mesh understanding and generation MLLM that natively processes text, images, and 3D geometry within a single Mixture-of-Transformers sequence stream. This section formalizes the sparse-voxel-based mesh tokenization, the MoT backbone with decoupled understanding and generation experts, and the conditional flow matching formulation that enables semantic-geometric alignment and context-aware generation.

\subsection{Architecture: Unified 3D Multimodal Mixture-of-Transformers}

EVA01 extends a pre-trained MLLM backbone (Qwen3-VL~\cite{Qwen3-VL}) to model 3D mesh geometry as a conditional flow matching problem within a unified sequence stream. Given a multimodal input sequence $\mathbf{X} = [\mathbf{x}_{\text{txt}}, \mathbf{x}_{\text{img}}, \mathbf{x}_{\text{mesh}}]$, where $\mathbf{x}_{(\cdot)}$ denotes the tokenized sequence of each modality, we learn the conditional probability distribution $p(\mathbf{x}_{\text{mesh}} \mid \mathbf{x}_{\text{txt}}, \mathbf{x}_{\text{img}}, \mathbf{c}_{\text{ctx}})$, with $\mathbf{c}_{\text{ctx}}$ encoding the historical context in multi-turn interactions.

\textbf{3D Mesh Tokenization.}
Following TRELLIS.2~\cite{xiang2025native}, we represent 3D assets as sparse voxel-based structures (O-Voxel) that jointly encode geometry and appearance.
Each asset is defined as a collection of feature tuples anchored to a regular 3D grid of resolution $N^3$:
\begin{equation}
    \boldsymbol{f} = \{(\boldsymbol{f}^{\text{shape}}_i, \boldsymbol{f}^{\text{mat}}_i, \boldsymbol{p}_i)\}_{i=1}^{L},
    \label{eq:ovoxel}
\end{equation}
where $\boldsymbol{f}^{\text{shape}}_i$ encodes local geometric information (dual vertex position, edge intersection flags, and splitting weights), $\boldsymbol{f}^{\text{mat}}_i$ encodes PBR material parameters (base color, metallic, roughness, opacity), and $\boldsymbol{p}_i \in \{0,\ldots,N-1\}^3$ is the coordinate of the $i$-th active voxel; inactive voxels are discarded.
This representation supports direct bidirectional conversion to and from meshes via a Flexible Dual Grid formulation, avoiding the iterative decoding and field extraction of prior representations.
A pre-trained VAE compresses the O-Voxel representation in Eq.~\ref{eq:ovoxel} into compact sparse latent tokens $\mathbf{x}_{\text{mesh}} \in \mathbb{R}^{L \times C}$, where $L$ is the number of active voxels.

\begin{table}[t]
    \centering
    \small
    \setlength{\tabcolsep}{2pt}
    \renewcommand{\arraystretch}{1.18}

    \caption{\textbf{Unified Block Attention Masking for Multi-Turn Editing.} Visibility constraints within a packed sequence. \colorbox{c_causal}{Purple}: Causal; \colorbox{c_bidir}{Green}: Bidirectional; \colorbox{c_mask}{Light gray}: Masked. The staged 3D latent blocks consist of sparse structure (SS; dense latent), sparse shape (shape; sparse latent), and sparse material (material; sparse latent). The current generation conditions on clean historical geometry while noisy blocks are hidden from all later blocks.}
    \label{tab:attn_mask}
    
    \resizebox{\linewidth}{!}{%
    \arrayrulecolor{eva_purple}
    \begin{tabular}{l|c|c|c|c|c|c|c|c|c|c}
        \hline
        \rowcolor{c_header}
        \textcolor{white}{\textbf{Query} $\downarrow$ \textbf{Key} $\rightarrow$} &
        \textcolor{white}{\textbf{Text 1}} &
        \textcolor{white}{\textbf{Image 1}} &
        \textcolor{white}{\textbf{Noise SS 1}} &
        \textcolor{white}{\textbf{Clean SS 1}} &
        \textcolor{white}{\textbf{Noise Shape 1}} &
        \textcolor{white}{\textbf{Clean Shape 1}} &
        \textcolor{white}{\textbf{Noise Material 1}} &
        \textcolor{white}{\textbf{Clean Material 1}} &
        \textcolor{white}{\textbf{Text 2}} &
        \textcolor{white}{\textbf{Noisy SS 2}} \\
        \hline
        
        \textbf{Text 1} &
        \cellcolor{c_causal}Causal & \cellcolor{c_mask}- & \cellcolor{c_mask}- & \cellcolor{c_mask}- & \cellcolor{c_mask}- & \cellcolor{c_mask}- & \cellcolor{c_mask}- & \cellcolor{c_mask}- & \cellcolor{c_mask}- & \cellcolor{c_mask}- \\
        \hline
        
        \textbf{Image 1} &
        \cellcolor{c_bidir}Full & \cellcolor{c_causal}Causal & \cellcolor{c_mask}- & \cellcolor{c_mask}- & \cellcolor{c_mask}- & \cellcolor{c_mask}- & \cellcolor{c_mask}- & \cellcolor{c_mask}- & \cellcolor{c_mask}- & \cellcolor{c_mask}- \\
        \hline
        
        \textbf{Noise SS 1} &
        \cellcolor{c_bidir}Full & \cellcolor{c_bidir}Full & \cellcolor{c_bidir}\textbf{Bi-Dir} & \cellcolor{c_mask}- & \cellcolor{c_mask}- & \cellcolor{c_mask}- & \cellcolor{c_mask}- & \cellcolor{c_mask}- & \cellcolor{c_mask}- & \cellcolor{c_mask}- \\
        \hline
        
        \textbf{Clean SS 1} &
        \cellcolor{c_bidir}Full & \cellcolor{c_bidir}Full & \cellcolor{c_mask}- & \cellcolor{c_bidir}\textbf{Bi-Dir} & \cellcolor{c_mask}- & \cellcolor{c_mask}- & \cellcolor{c_mask}- & \cellcolor{c_mask}- & \cellcolor{c_mask}- & \cellcolor{c_mask}- \\
        \hline
        
        \textbf{Noise Shape 1} &
        \cellcolor{c_bidir}Full & \cellcolor{c_bidir}Full & \cellcolor{c_mask}- & \cellcolor{c_bidir}Full & \cellcolor{c_bidir}\textbf{Bi-Dir} & \cellcolor{c_mask}- & \cellcolor{c_mask}- & \cellcolor{c_mask}- & \cellcolor{c_mask}- & \cellcolor{c_mask}- \\
        \hline
        
        \textbf{Clean Shape 1} &
        \cellcolor{c_bidir}Full & \cellcolor{c_bidir}Full & \cellcolor{c_mask}- & \cellcolor{c_bidir}Full & \cellcolor{c_mask}- & \cellcolor{c_bidir}\textbf{Bi-Dir} & \cellcolor{c_mask}- & \cellcolor{c_mask}- & \cellcolor{c_mask}- & \cellcolor{c_mask}- \\
        \hline

        \textbf{Noise Material 1} &
        \cellcolor{c_bidir}Full & \cellcolor{c_bidir}Full & \cellcolor{c_mask}- & \cellcolor{c_bidir}Full & \cellcolor{c_mask}- & \cellcolor{c_bidir}Full & \cellcolor{c_bidir}\textbf{Bi-Dir} & \cellcolor{c_mask}- & \cellcolor{c_mask}- & \cellcolor{c_mask}- \\
        \hline
        
        \textbf{Clean Material 1} &
        \cellcolor{c_bidir}Full & \cellcolor{c_bidir}Full & \cellcolor{c_mask}- & \cellcolor{c_bidir}Full & \cellcolor{c_mask}- & \cellcolor{c_bidir}Full & \cellcolor{c_mask}- & \cellcolor{c_bidir}\textbf{Bi-Dir} & \cellcolor{c_mask}- & \cellcolor{c_mask}- \\
        \hline
        
        \textbf{Text 2} &
        \cellcolor{c_bidir}Full & \cellcolor{c_bidir}Full & \cellcolor{c_mask}- & \cellcolor{c_bidir}Full & \cellcolor{c_mask}- & \cellcolor{c_bidir}Full & \cellcolor{c_mask}- & \cellcolor{c_bidir}Full & \cellcolor{c_causal}Causal & \cellcolor{c_mask}- \\
        \hline
        
        \textbf{Noisy SS 2} &
        \cellcolor{c_bidir}Full & \cellcolor{c_bidir}Full & \cellcolor{c_mask}- & \cellcolor{c_bidir}Full & \cellcolor{c_mask}- & \cellcolor{c_bidir}Full & \cellcolor{c_mask}- & \cellcolor{c_bidir}Full & \cellcolor{c_bidir}Full & \cellcolor{c_bidir}\textbf{Bi-Dir} \\
        \hline
    \end{tabular}
    \arrayrulecolor{black}
    }
    \vspace{-6pt}
\end{table}

EVA01 adopts a structured sparse grid representation in place of the VecSet paradigm~\cite{zhang2023shapetovecset} common in recent 3D generative frameworks.
VecSet compresses point cloud data into order-invariant latent tokens via a Perceiver-like~\cite{jaegle2021perceiver} architecture; while compact, these unordered tokens lack explicit spatial coordinates, depriving attention mechanisms of the positional grounding required to establish stable geometric correspondences.
Our sparse voxel representation anchors each token to a fixed coordinate $\boldsymbol{p}_i$, binding geometry to a regular spatial lattice---a structural regularity indispensable for long-context interleaved sequences, where sequence position and spatial topology must be tracked simultaneously.
The sparse convolutional backbone further enables mesh sampling at substantially higher resolutions than VecSet methods, yielding reconstruction precision well beyond prior approaches.
Ablation experiments~(Sec.~\ref{subsec:ablation}) confirm that explicit coordinate injection is a prerequisite for preventing geometric collapse in native generation.

\textbf{Mixture-of-Transformers Backbone.}
To resolve the optimization dichotomy between semantic reasoning and geometric generation, we construct a multimodal MoT architecture~\cite{liang2025mixtureoftransformers} derived from the Qwen3-VL~\cite{Qwen3-VL} backbone. We strictly decouple model parameters---spanning FFNs and attention projections---into two specialized sets: an Understanding Expert ($E_{\text{und}}$) and a Generation Expert ($E_{\text{gen}}$), as shown in Fig.~\ref{fig:architecture}.

\textit{Dual-Expert Routing.}
We employ a deterministic hard-routing strategy where computation for the $i$-th token is governed by its modality $m_i$ (with $\text{txt},\text{img} \mapsto E_{\text{und}}$ and $\text{mesh} \mapsto E_{\text{gen}}$). Formally, any linear transformation is computed via direct parameter indexing using modality-specific weight matrices $\mathbf{W}^{(m)}$:
\begin{equation}
    \mathbf{h}'_i = \mathbf{h}_i \mathbf{W}^{(m_i)}.
\end{equation}
This formulation segregates optimization: $E_{\text{und}}$ remains a stable semantic anchor preserving the pre-trained MLLM priors, while $E_{\text{gen}}$ undergoes geometric optimization without compromising understanding.

\textit{Understanding Expert Design.}
Following Janus~\cite{wu2025janus}, $E_{\text{und}}$ decouples modality-specific encoding from the shared semantic backbone. Text is tokenized via the Qwen tokenizer, images are encoded through SigLIP2~\cite{tschannen2025siglip2}, and 3D mesh geometry is processed by a pre-trained Point-BERT~\cite{xu2024pointllm} encoder. These modality-specific representations are projected into a shared semantic space derived from Qwen3-VL~\cite{Qwen3-VL}, whose pre-aligned multimodal feature manifold exhibits 3D spatial understanding. This design preserves $E_{\text{und}}$ as a semantic anchor without requiring modality-specific fine-tuning of the encoder stacks.

\textit{Generation Expert Design.}
$E_{\text{gen}}$ implements a three-stage flow matching pipeline that mirrors the hierarchical structure of the sparse voxel representation. The \textbf{sparse structure} stage operates on a dense latent grid and predicts the occupancy layout---which voxels are active---via a decoder with progressive upsampling, yielding the coordinate scaffolding for subsequent stages. Conditioned on this sparse layout, the \textbf{sparse geometry} stage generates shape features $\boldsymbol{f}^{\text{shape}}$ within active voxels using sparse convolutions and attention restricted to occupied coordinates. Finally, the \textbf{sparse material} stage synthesizes PBR material features $\boldsymbol{f}^{\text{mat}}$ aligned to the generated geometry, similarly employing sparse operators. This ensures computational cost for the latter two stages scales with surface area rather than volume. At inference, the stages execute sequentially (structure $\rightarrow$ geometry $\rightarrow$ material); during training, they are parallelized via the unified attention mask (Table~\ref{tab:attn_mask}).

\textit{Shared Global Attention.}
To enable cross-modal reasoning despite parameter isolation, we implement shared global attention. For a token $i$ with modality $m_i$, the output $\mathbf{y}_i$ is computed as:
\begin{align}
    \mathbf{q}_i, \mathbf{k}_i, \mathbf{v}_i &= \mathbf{h}_i \mathbf{W}_Q^{(m_i)}, \mathbf{h}_i \mathbf{W}_K^{(m_i)}, \mathbf{h}_i \mathbf{W}_V^{(m_i)} \\
    \mathbf{y}_i &= \text{Attn}(\mathbf{q}_i, \mathbf{K}, \mathbf{V}; \mathbf{M}) \mathbf{W}_O^{(m_i)}
\end{align}
where $\mathbf{K}, \mathbf{V}$ aggregate keys and values from all sequence tokens $j$ using their respective expert weights $\mathbf{W}_{\{K,V\}}^{(m_j)}$. The unified mask $\mathbf{M}$ (Table~\ref{tab:attn_mask}) controls information flow, enabling the generation expert to query semantic priors from the understanding expert.

\textit{3D MRoPE.}
Standard 1D positional encodings flatten geometric structures, disrupting volumetric topology. Following Qwen3-VL~\cite{Qwen3-VL}, we propose a 3D Interleaved MRoPE strategy that repurposes the original $(T, W, H)$ rotary embeddings for sparse grid coordinates $(x, y, z)$ in the generation branch. Rotary frequencies for the $X$, $Y$, and $Z$ axes are interleaved across the feature vector to maximize cross-dimension interaction, decomposing the head dimension $D$ into strided subspaces indexed by $\mathcal{I}_x, \mathcal{I}_y, \mathcal{I}_z$ (e.g., $\mathcal{I}_x = \{k \mid k \equiv 0 \pmod 3\}$). For a mesh token at sparse voxel coordinate $\mathbf{p}=(x, y, z)$, the rotary embedding is applied as:
\begin{equation}
    \text{RoPE}(\mathbf{x}, \mathbf{p}) = \text{Interleave}\left( \mathcal{R}_x(\mathbf{x}_{\mathcal{I}_x}), \mathcal{R}_y(\mathbf{x}_{\mathcal{I}_y}), \mathcal{R}_z(\mathbf{x}_{\mathcal{I}_z}) \right),
\end{equation}
where $\mathcal{R}_{\phi}$ denotes rotation by spatial frequency $\phi$, and $\mathbf{x}_{\mathcal{I}}$ represents the feature slice corresponding to subspace indices. This distributes spatial information uniformly across the sparse grid, injecting Euclidean inductive biases that prevent geometric drift during long-context editing.

\subsection{Data: From Static Assets to Contextual Editing}
\label{sec:data}

To resolve the scarcity of paired 3D-text data and enable context-aware editing, we curate a hierarchical dataset in two phases: first assembling a large-scale corpus of high-fidelity static 3D assets, then synthesizing multi-turn interleaved editing trajectories from this foundation.

\textbf{Phase 1: High-Quality Static 3D Asset Curation.}
We aggregate approximately 1.2M raw 3D assets from Objaverse-XL~\cite{deitke2023objaverse_xl}, TexVerse~\cite{zhang2025texverse}, and licensed internal repositories, spanning household objects, vehicles, characters, and mechanical assemblies.
A rigorous standardization pipeline processes the full corpus while identifying a premium subset of 400K assets through stricter quality criteria:
\textit{1) Geometric Standardization:} We canonicalize mesh orientations to a consistent world frame, verify UV coordinate integrity, validate texture resolution, and repair non-manifold edges and degenerate faces. Assets with irrecoverable geometric defects are discarded.
\textit{2) Aesthetic and Structural Filtering:} A learned aesthetic scoring model trained on human preference annotations assigns quality ratings based on geometric complexity, texture realism, and visual appeal. The top 400K assets form a high-quality subset reserved for later training stages, while the full 1.2M corpus is retained for foundational pretraining. Topological analysis further tags objects with distinct structural properties---rigid sub-meshes amenable to part-based editing, articulated components with defined joint hierarchies, and assets with skeletal animation data---enabling targeted downstream editing tasks.
\textit{3) Multi-View Dense Captioning:} Each asset is rendered from 8--12 uniformly sampled camera viewpoints under varying HDR environment lighting. A VLM processes these renderings to produce captions at three granularities: a short category-level summary, a medium-length description of geometric structure and material properties, and a detailed paragraph-level caption covering fine-grained shape details, surface texture, and functional affordances. These multi-granularity captions serve distinct training objectives across curriculum stages.
The resulting dataset is formalized as triplets $\mathcal{D}_{\text{static}} = \{ (\mathbf{t}_i, \{\mathbf{I}_{i,m}\}_{m=1}^M, \mathbf{x}_{\text{mesh}, i}) \}$, where $\mathbf{t}_i$ is the caption, $\{\mathbf{I}_{i,m}\}$ the multi-view renders, and $\mathbf{x}_{\text{mesh}, i}$ denotes the sparse voxel latent tokens comprising sparse structure, geometry, and material latent sets.

\begin{figure}[t]
  \centering
  \includegraphics[width=1\linewidth]{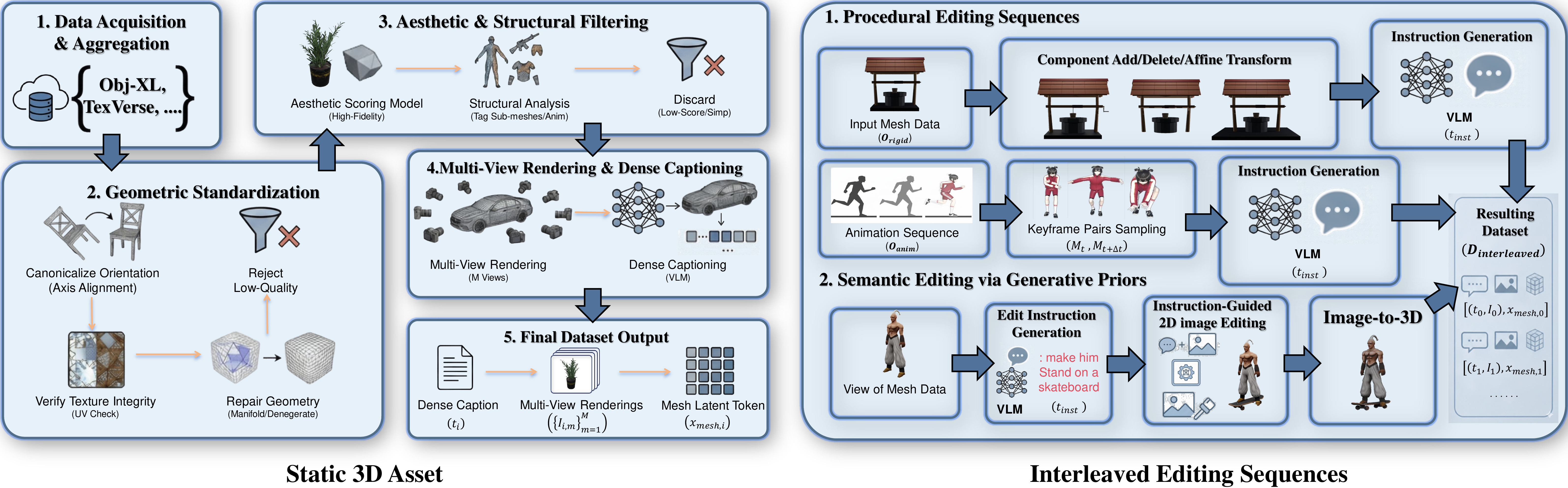}
  \caption{
  \textbf{Data Curation Pipeline of EVA01.}
  \textbf{(Left) Static 3D Asset Curation:} We standardize raw 3D assets through geometric canonicalization, aesthetic filtering, and multi-view dense captioning to construct high-quality text-image-mesh triplets.
  \textbf{(Right) Interleaved Editing Sequences:} To enable context-aware editing, we synthesize multi-turn sequences via two complementary pathways: \textit{Procedural Editing} (top right) utilizing rigid transformations and animation keyframes for structural precision, and \textit{Semantic Editing} (bottom right) leveraging 2D generative priors for open-ended stylistic modification.
  }
  \label{fig:data}
\end{figure}

\textbf{Phase 2: Interleaved Editing Sequences.}
We synthesize large-scale multi-turn editing sequences via two complementary pipelines, generating approximately 3M procedural and 300K semantic editing trajectories from the curated static asset pool.
\textit{Procedural Editing:} We algorithmically generate editing sequences using a composite operation set $\mathcal{O}_{\text{proc}} = \mathcal{O}_{\text{rigid}} \cup \mathcal{O}_{\text{anim}} \cup \mathcal{O}_{\text{topo}}$. $\mathcal{O}_{\text{rigid}}$ spans 6-DoF affine transformations (translation, rotation, non-uniform scaling) applied to individual sub-mesh components, enabling part rearrangement, structural resizing, and component duplication. $\mathcal{O}_{\text{topo}}$ introduces topological perturbations including mesh boolean operations (union, difference, intersection) and localized mesh deformation fields, modeling constructive and destructive editing actions (e.g., adding a handle, carving a cavity). $\mathcal{O}_{\text{anim}}$ derives editing pairs from continuous animation sequences---skeletal deformations, mechanical articulations, and blend-shape morphs---by sampling frame pairs $(M_t, M_{t+\Delta t})$ with variable temporal stride $\Delta t$. Larger strides encourage learning long-range kinematic chains, while smaller strides provide fine-grained deformation supervision.
\textit{Semantic Editing:} An LLM generates diverse editing instructions $\mathbf{t}_{\text{inst}}$ for each source mesh $M_0$, covering attribute changes (material, color, texture), structural modifications (shape deformation, part replacement), and stylistic transformations. A conditioned image editing model modifies a representative view $\mathbf{I}_0$ to $\mathbf{I}'$ following $\mathbf{t}_{\text{inst}}$, which is then lifted to 3D via a reconstruction pipeline with multi-view consistency enforcement. This pathway transfers semantic priors from 2D generative models into the 3D editing domain.
The interleaved dataset is formalized as $\mathcal{D}_{\text{interleaved}} = \{\mathcal{S}^{(k)}\}$, where each sequence $\mathcal{S} = [(\mathbf{t}_0, \mathbf{I}_0), \mathbf{x}_{\text{mesh}, 0}, (\mathbf{t}_{\text{inst}, 1}, \mathbf{I}_{\text{ref}, 1}), \mathbf{x}_{\text{mesh}, 1}, \dots, (\mathbf{t}_{\text{inst}, T}, \mathbf{I}_{\text{ref}, T}), \mathbf{x}_{\text{mesh}, T}]$ spans $T$ editing turns, with each turn conditioning on the accumulated geometric and semantic context.
Figure~\ref{fig:data} summarizes this static-to-interleaved data construction pipeline.

\subsection{Training: Multi-Stage Curriculum Learning}

To bridge the misalignment between textual semantics and 3D topology, we employ a curriculum learning strategy that progressively aligns modalities and enables increasingly complex capabilities.
The training objective combines two losses. For 3D generation, we formulate mesh synthesis as a Conditional Flow Matching (CFM) problem~\cite{lipman2023flow}. Let $\mathbf{x}_1 \sim q(\mathbf{x}_{\text{mesh}})$ denote the clean sparse voxel latent and $\mathbf{x}_0 \sim \mathcal{N}(\mathbf{0}, \mathbf{I})$ standard Gaussian noise. Defining the probability path $\mathbf{x}_t = (1 - t)\mathbf{x}_0 + t\mathbf{x}_1$, the flow matching loss is:
\begin{equation}
    \mathcal{L}_{\text{FM}}(\theta) = \mathbb{E}_{t, \mathbf{x}_0, \mathbf{x}_1, \mathbf{c}} \left[ \left\| \mathbf{v}_{\theta}(\mathbf{x}_t, t, \mathbf{c}) - (\mathbf{x}_1 - \mathbf{x}_0) \right\|^2 \right],
\end{equation}
where $\mathbf{c}$ denotes the conditional context. For mesh understanding, we employ a standard autoregressive cross-entropy loss over text tokens conditioned on mesh features:
\begin{equation}
    \mathcal{L}_{\text{CE}}(\theta) = -\sum_{i=1}^{T} \log p_{\theta}(t_i \mid t_{<i}, \mathbf{x}_{\text{mesh}}),
\end{equation}
where $\{t_i\}$ are text caption tokens and $\mathbf{x}_{\text{mesh}}$ is the sparse voxel latent of the corresponding 3D asset. Optimization proceeds through five stages, each introducing a distinct capability while preserving previously acquired knowledge.

\textbf{Stage 1: Mesh Understanding Warm-up.}
We begin by establishing mesh understanding capability. Using mesh-text paired data from $\mathcal{D}_{\text{static}}$, we train only a lightweight MLP projector that maps Point-BERT~\cite{xu2024pointllm} mesh features into the Qwen3-VL~\cite{Qwen3-VL} text embedding space. All other parameters---including $E_{\text{und}}$, $E_{\text{gen}}$, the mesh VAE, and the modality-specific encoders---remain frozen. This stage is optimized solely with $\mathcal{L}_{\text{CE}}$ on the captioning objective, warming up the understanding pathway at minimal cost while preserving the pre-trained MLLM backbone.

\textbf{Stage 2: Visual-Geometric Initialization.}
With a functioning understanding pathway in place, we establish generation capability by mapping dense visual features to the 3D manifold. Qwen3-VL's DeepStack strategy~\cite{Qwen3-VL} propagates low-level visual features from early SigLIP2~\cite{tschannen2025siglip2} layers directly into the LLM backbone, providing dense spatial and textural cues essential for accurate image-to-3D reconstruction. Ablation experiments~(Sec.~\ref{subsec:ablation}) confirm that removing this pathway significantly degrades geometric fidelity.
We train $E_{\text{gen}}$ with $\mathcal{L}_{\text{FM}}$ using image-conditioned samples from $\mathcal{D}_{\text{static}}$, while $E_{\text{und}}$ is jointly trained with mesh-text paired data under $\mathcal{L}_{\text{CE}}$. This dual-objective setup creates a synergistic effect: the mesh-text understanding signal sharpens $E_{\text{und}}$'s geometric awareness, and through shared global self-attention, these refined semantic representations directly inform $E_{\text{gen}}$'s image-to-3D generation, improving reconstruction fidelity beyond what image conditioning alone can achieve. A reduced learning rate on $E_{\text{und}}$ preserves its pre-trained multimodal priors while enabling this cross-task transfer.

\textbf{Stage 3: Semantic Modality Alignment.}
To bridge the textual and geometric manifolds, we introduce \textit{Triple-Batch Sampling} combined with \textit{Modality Dropout}. For each triplet in $\mathcal{D}_{\text{static}}$, we construct three independent training samples conditioned on text, images, and mesh-text pairs respectively. Dynamic image token dropout ($p_{\text{drop}}$) compels the network to reconstruct $\mathbf{x}_{\text{mesh}}$ from textual cues alone, implicitly distilling the visual-geometric priors from Stage~2 into the text-conditioned generation process. Mesh-text pairing continues to reinforce understanding via $\mathcal{L}_{\text{CE}}$, while $E_{\text{und}}$ and $E_{\text{gen}}$ are jointly optimized with modality-specific learning rates to ensure output consistency across conditioning modalities.

\textbf{Stage 4: Context-Aware Instruction Tuning.}
We enable sequential, stateful 3D editing using the interleaved dataset $\mathcal{D}_{\text{interleaved}}$. The generation of the $k$-th geometric state is conditioned on the full interaction history $\mathbf{c}_k = \{ \mathbf{t}_{\text{inst}}, \mathbf{x}_{\text{hist}} \}$, with 3D Interleaved MRoPE encoding both global interaction timestamps and local spatial coordinates. The attention mask (Table~\ref{tab:attn_mask}) enforces that $\mathbf{x}_k$ attends to the clean latent of $\mathbf{x}_{k-1}$, enabling differential editing operations---modifying topology based on instructions while preserving identity across turns---rather than stateless regeneration. This stage employs both $\mathcal{L}_{\text{FM}}$ and $\mathcal{L}_{\text{CE}}$ to maintain understanding quality alongside the emerging editing capability.

\textbf{Stage 5: High-Quality Finetuning.}
The final stage refines generation fidelity on the curated 400K high-quality asset subset. Training continues with $\mathcal{L}_{\text{FM}}$ at reduced learning rates, sharpening geometric details and surface quality. This stage is critical for elevating the upper bound of visual fidelity while preserving the context-aware editing behavior and mesh understanding capabilities acquired in prior stages.

\begin{table}[tb]
    \centering
    \small
    \caption{\textbf{Training Recipe of EVA01.} Multi-stage curriculum with differential optimization. $E_{\text{und}}$: Understanding Expert; $E_{\text{gen}}$: Generation Expert. MSE denotes the flow-matching regression loss. \textcolor{eva_green}{Green highlight}: interleaved editing data.}
    \label{tab:training_recipe}

    \setlength{\tabcolsep}{4pt}
    \setlength{\aboverulesep}{0.25ex}
    \setlength{\belowrulesep}{0.25ex}
    \renewcommand{\arraystretch}{1.02}

    \arrayrulecolor{eva_purple}
    \begin{tabular}{@{}>{\raggedright\arraybackslash}p{0.20\linewidth}|*{5}{>{\centering\arraybackslash}p{0.104\linewidth}}@{}}
        \toprule
        \rowcolor{eva_purple}
        & \textcolor{white}{\shortstack[c]{\textbf{Stage 1}\\[-1pt]\textbf{(Warm-up)}}} & \textcolor{white}{\shortstack[c]{\textbf{Stage 2}\\[-1pt]\textbf{(Init)}}} & \textcolor{white}{\shortstack[c]{\textbf{Stage 3}\\[-1pt]\textbf{(Align)}}} & \textcolor{white}{\shortstack[c]{\textbf{Stage 4}\\[-1pt]\textbf{(Edit)}}} & \textcolor{white}{\shortstack[c]{\textbf{Stage 5}\\[-1pt]\textbf{(HQ FT)}}} \\
        \midrule
        \rowcolor{eva_lavender}
        \multicolumn{6}{l}{\textit{\textbf{Hyperparameters}}} \\
        Learning rate ($E_{\text{gen}}$) & \textbf{0.0} & $2\text{e-}4$ & $1\text{e-}4$ & $5\text{e-}5$ & $5\text{e-}6$ \\
        Learning rate ($E_{\text{und}}$) & $1\text{e-}4$ & $1\text{e-}5$ & $1\text{e-}5$ & $5\text{e-}6$ & $1\text{e-}6$ \\
        LR scheduler & Constant & Constant & Cosine & Cosine & Constant \\
        Optimizer & \multicolumn{5}{c}{AdamW ($\beta_{1,2}{=}0.9, .95; \epsilon{=}10^{-8}; \text{wd}{=}0.05$)} \\
        Training steps & 20K & 150K & 100K & 80K & 30K \\
        Sequence length & 32-40K & 56-64K & 56-64K & 80-96K & 80-96K \\
        Max context window & 24K & 30K & 30K & 60K & 60K \\
        Mesh resolution & -- & $512^3$ & $512^3$ & $512^3$ & $512^3$ \\
        Diffusion timestep shift & -- & 1.0 & 3.0 & 3.0 & 3.0 \\
        Loss weight (CE : MSE) & 1:0 & 0.25:1 & 0.25:1 & 0.25:1 & 0:1 \\
        \midrule
        \rowcolor{eva_lavender}
        \multicolumn{6}{l}{\textit{\textbf{Data Sampling Ratio}}} \\
        Image-Text Pair (I2T) & 0.0 & 0.0 & 0.05 & 0.05 & 0.0 \\
        Image-Mesh Pair (I2M) & 0.0 & 0.9 & 0.3 & 0.1 & 0.1 \\
        Text-Mesh Pair (T2M) & 0.0 & 0.0 & 0.5 & 0.1 & 0.4 \\
        Text-Mesh Pair (M2T) & 1.0 & 0.1 & 0.15 & 0.1 & 0.0 \\
        \rowcolor{eva_mint}
        Procedural (Interleaved) & 0.0 & 0.0 & 0.0 & 0.35 & 0.1 \\
        \rowcolor{eva_mint}
        Semantic (Interleaved) & 0.0 & 0.0 & 0.0 & 0.3 & 0.4 \\
        \bottomrule
    \end{tabular}%
    \arrayrulecolor{black}
\end{table}

\section{Experiments}

We evaluate EVA01 across four capability axes: single-turn generation (text-to-3D and image-to-3D), multi-turn context-aware editing, and mesh understanding. Beyond standard benchmarks, we report ablation studies dissecting key architectural and training components, analyze learned representations, and derive insights from training dynamics.

\subsection{Experimental Protocol}

\paragraph{Data.}
We follow the data curation pipeline described in Section~\ref{sec:data}, aggregating 1.2M raw assets (filtered to a 400K premium subset) for static training and synthesizing 3M procedural and 300K semantic editing sequences for multi-turn interactivity. Per-stage data sampling ratios are specified in Table~\ref{tab:training_recipe}.

For the understanding task, we build on the released PointLLM-V2 raw\_v2 corpus, a point-native object-centric 3D-language dataset. We use its stage-structured split, namely Stage1-1M for point-language alignment and Stage2-700k for instruction tuning. This organization is well matched to our design, enabling disentanglement of faithful 3D semantic grounding from higher-level instruction-following behavior.
In addition, we leverage our curated text-mesh pairs for caption training and employ the interleaved editing dataset to predict semantic editing instructions, making full use of our constructed data.

\paragraph{Implementation.}
EVA01 is built upon \textbf{Qwen3-VL-2B-Instruct}~\cite{Qwen3-VL} as the MLLM backbone. The Generation Expert ($E_{\text{gen}}$) is structurally mirrored from $E_{\text{und}}$ and initialized from its pre-trained weights to accelerate convergence. The pre-trained mesh VAE and Point-BERT~\cite{xu2024pointllm} encoder remain frozen throughout all training stages; the visual encoder SigLIP2~\cite{tschannen2025siglip2} is trained.
Training uses AdamW~\cite{loshchilovdecoupled} ($\beta_1{=}0.9$, $\beta_2{=}0.95$, weight decay${=}0.05$) with bfloat16 mixed precision and FlexAttention. A cosine learning rate schedule with linear warmup (1,000--5,000 steps, stage-dependent) is employed; per-stage learning rates and training steps are detailed in Table~\ref{tab:training_recipe}. Training uses sample packing with a batch size of 1---multiple samples concatenated into a single packed sequence---distributed across 32 NVIDIA H20 (96GB) GPUs via Fully Sharded Data Parallel (FSDP). Classifier-free guidance with a conditioning dropout rate of $0.1$ is applied across all flow-matching stages. Total training wall-clock time is approximately two months.

\paragraph{Evaluation Benchmarks and Metrics.}
We evaluate across three axes.
For \textbf{single-turn generation}, we use the standard \textbf{Toys4K}~\cite{stojanov21cvpr} benchmark (3,218 high-quality 3D assets), reporting CLIP Score~\cite{hessel2021clipscore} for text-shape semantic alignment, and Fr\'{e}chet Distance (FD) and Kernel Distance (KD) based on DINOv2~\cite{oquab2023dinov2} features for geometric and visual fidelity.
For \textbf{multi-turn editing}, we curate a 400-sample evaluation set from our interleaved editing corpus, selecting 200 procedural and 200 semantic editing sequences. Existing local-editing benchmarks such as Edit3D-Bench are useful references, but their limited scale (roughly 100 source objects), single-turn structure, and coarse edit-region coverage do not fully stress long-context mesh-native editing. For precise geometric edits, we provide manually verified masks to support mask-dependent baselines; for broad style and semantic edits, the masks indicate only approximate affected regions. EVA01 itself does not consume masks at inference time. We evaluate unedited-region consistency using Chamfer Distance (CD) and masked multi-view PSNR, and assess overall editing quality with CLIP, FD$_{\text{DINOv2}}$, and user-study preference (Pref\%).

For \textbf{mesh understanding}, we evaluate on the PointLLM-200 Objaverse captioning benchmark~\cite{xu2024pointllm}, which contains 200 held-out Objaverse objects with official PointLLM reference captions from the human-annotated Cap3D split. We use the prompt \emph{``Caption this 3D model in detail.''} under a prompt-only generation protocol: the model receives only the 3D input and the captioning prompt, while the reference caption is used solely for scoring. This avoids teacher-forcing or continuation-style leakage where the reference answer is included in the model context.

We report lexical, semantic, and judge-based captioning metrics. BLEU-$n$~\cite{papineni2002bleu}, ROUGE-L~\cite{lin2004rouge}, and METEOR~\cite{banerjee2005meteor} are computed with standard implementations and reported on a $0$--$100$ scale. Semantic agreement beyond token overlap is measured via cosine similarity between reference and generated captions using Sentence-BERT~\cite{reimers2019sentencebert} and SimCSE~\cite{gao2021simcse}, scaled by $100$. We further report two complementary GPT-based judge metrics. \textbf{GPT-ref} follows the PointLLM protocol~\cite{xu2024pointllm}: the judge is given the reference caption and the model prediction, estimates the fraction of reference aspects correctly or partially covered by the prediction, and returns a score in $[0,100]$ averaged over all valid responses. Since PointLLM-200 references are often brief single-sentence captions, we additionally report \textbf{GPT-img}, a render-grounded judge score. For GPT-img, GPT-5.5 is given four RGB renders of the object (front, right, back, and left) together with the candidate caption, but no reference caption, and scores how faithfully the caption describes the visible 3D object. All metrics are computed with the same scorer within each metric family, and higher is better.

\textbf{Baselines.}
For single-turn generation, we compare against Shap-E~\cite{jun2023shap}, 3DTopia-XL~\cite{chen20253dtopia}, TRELLIS~\cite{xiang2024structured}, Michelangelo~\cite{zhao2023michelangelo}, 3DGen-R1~\cite{tang2025we}, GVGEN~\cite{he2024gvgen}, TRELLIS.2~\cite{xiang2025native}, Hunyuan3D-2.1~\cite{hunyuan3d2025hunyuan3d}, Direct3D-S2~\cite{wu2026direct3d}, Step1X-3D~\cite{li2025step1x}, Hi3DGen~\cite{ye2025hi3dgen}, and ShapeLLM-Omni~\cite{ye2025shapellm}.
For multi-turn editing, we compare against Instant3DiT~\cite{barda2025instant3dit}, TRELLIS~\cite{xiang2024structured}, and VoxHammer~\cite{li2025voxhammer}.
For mesh understanding, we compare against PointLLM-7B/13B~\cite{xu2024pointllm}, ShapeLLM-7B/13B~\cite{qi2024shapellm}, MiniGPT-3D~\cite{tang2024minigpt3d}, LLaMA-Mesh~\cite{wang2024llamamesh}, GreenPLM~\cite{tang2025more}, and ShapeLLM-Omni~\cite{ye2025shapellm}.
All baselines are evaluated using their released checkpoints and official inference protocols.
\begin{table}[tb]
    \centering
    \setlength{\tabcolsep}{2pt}
    \renewcommand{\arraystretch}{1.12}
    \caption{\textbf{Quantitative comparisons on Toys4K.} We report CLIP, FD$_{\text{DINOv2}}$, KD$_{\text{DINOv2}}$, and user-study preference (Pref). KD is reported $\times 100$. \textcolor{gray}{N/A} denotes unsupported modalities.}
    \label{tab:toys4k_3col}
    \arrayrulecolor{eva_purple}
    \newcommand{\nacell}{\textcolor{gray}{N/A}}
        \resizebox{\linewidth}{!}{%
        \begin{tabular}{@{}>{\raggedright\arraybackslash}p{0.21\linewidth}|*{8}{>{\centering\arraybackslash}p{0.09\linewidth}}@{}}
        \toprule
        \rowcolor{eva_purple}
        \textcolor{white}{\textbf{Method}}
        & \multicolumn{4}{c|}{\textcolor{white}{\textbf{Text-to-3D}}}
        & \multicolumn{4}{c}{\textcolor{white}{\textbf{Image-to-3D}}} \\
        \rowcolor{eva_purple}
        \textcolor{white}{}
        & \textcolor{white}{\shortstack[c]{CLIP\\$\uparrow$}}
        & \textcolor{white}{\shortstack[c]{FD$_{\text{DINOv2}}$\\$\downarrow$}}
        & \textcolor{white}{\shortstack[c]{KD$_{\text{DINOv2}}$\\$\downarrow$}}
        & \textcolor{white}{\shortstack[c]{Pref\\$\uparrow$}}
        & \textcolor{white}{\shortstack[c]{CLIP\\$\uparrow$}}
        & \textcolor{white}{\shortstack[c]{FD$_{\text{DINOv2}}$\\$\downarrow$}}
        & \textcolor{white}{\shortstack[c]{KD$_{\text{DINOv2}}$\\$\downarrow$}}
        & \textcolor{white}{\shortstack[c]{Pref\\$\uparrow$}} \\
        \midrule

        \rowcolor{eva_purple}
        \multicolumn{9}{l}{\textcolor{white}{\textbf{Text-to-3D and Image-to-3D}}} \\

        Shap-E~\cite{jun2023shap}
        & 25.12 & 495.23 & 49.85 & 0.0\%
        & 82.15 & 465.11 & 62.55 & 0.0\% \\

        3DTopia-XL~\cite{chen20253dtopia}
        & 25.85 & 425.12 & 35.23 & 0.0\%
        & 83.42 & 380.88 & 45.15 & 0.0\% \\

        TRELLIS~\cite{xiang2024structured}
        & \underline{30.80} & \underline{238.45} & \underline{4.25} & \underline{14.8\%}
        & 85.65 & 67.32 & 0.75 & 5.1\% \\

        ShapeLLM-Omni~\cite{ye2025shapellm}
        & 26.70 & 310.55 & 18.20 & 5.1\%
        & 84.50 & 145.32 & 12.45 & 0.0\% \\

        Michelangelo~\cite{zhao2023michelangelo}
        & 25.92 & 405.36 & 27.40 & 0.0\%
        & 83.20 & 360.45 & 42.20 & 0.0\% \\

        \rowcolor{eva_purple}
        \multicolumn{9}{l}{\textcolor{white}{\textbf{Text-to-3D only}}} \\

        3DGen-R1~\cite{tang2025we}
        & 29.35 & 263.72 & 6.85 & 7.9\%
        & \nacell & \nacell & \nacell & \nacell \\

        GVGEN~\cite{he2024gvgen}
        & 25.55 & 438.90 & 30.65 & 1.9\%
        & \nacell & \nacell & \nacell & \nacell \\

        \rowcolor{eva_purple}
        \multicolumn{9}{l}{\textcolor{white}{\textbf{Image-to-3D only}}} \\

        TRELLIS.2~\cite{xiang2025native}
        & \nacell & \nacell & \nacell & \nacell
        & \textbf{89.34} & \textbf{56.82} & \textbf{0.49} & \textbf{41.7\%} \\

        Hunyuan3D-2.1~\cite{hunyuan3d2025hunyuan3d}
        & \nacell & \nacell & \nacell & \nacell
        & 86.95 & 64.80 & 0.66 & 10.2\% \\

        Direct3D-S2~\cite{wu2026direct3d}
        & \nacell & \nacell & \nacell & \nacell
        & 86.05 & 72.25 & 0.88 & 6.9\% \\

        Step1X-3D~\cite{li2025step1x}
        & \nacell & \nacell & \nacell & \nacell
        & 86.60 & 66.10 & 0.72 & 7.9\% \\

        Hi3DGen~\cite{ye2025hi3dgen}
        & \nacell & \nacell & \nacell & \nacell
        & 86.48 & 65.70 & 0.70 & 6.0\% \\

        \midrule
        \rowcolor{eva_mint}
        \textbf{Ours}
        & \textbf{35.72} & \textbf{122.48} & \textbf{1.18} & \textbf{70.4\%}
        & \underline{87.28} & \underline{61.74} & \underline{0.63} & \underline{22.2\%} \\

        \bottomrule
        \end{tabular}%
        }%
    \arrayrulecolor{black}
\end{table}

\subsection{Single-Turn Generation}

\begin{figure}[!htbp]
  \centering
  \includegraphics[width=\linewidth,height=0.56\textheight,keepaspectratio]{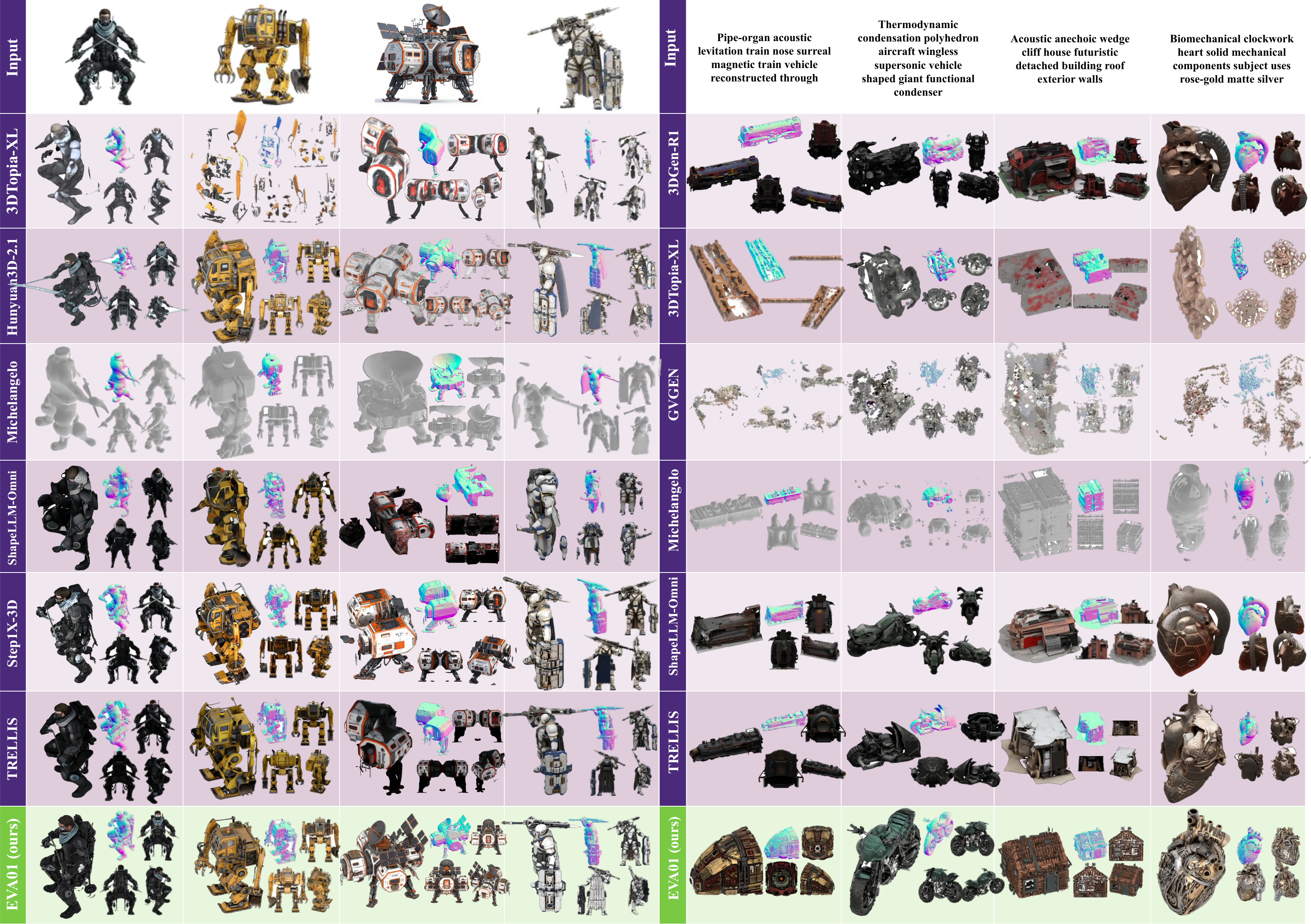}
  \caption{
  \textbf{Qualitative Comparison with Baselines.}
  We compare EVA01 with representative text-to-3D and image-to-3D baselines on Toys4K, including 3DTopia-XL, Hunyuan3D-2.1, Michelangelo, ShapeLLM-Omni, Step1X-3D, TRELLIS, 3DGen-R1, and GVGEN.
  Across both image-conditioned cases (left) and text-conditioned cases (right), EVA01 better preserves object-level semantics, part structure, and material consistency, producing complete meshes with coherent geometry where prior methods often suffer from missing components, distorted topology, over-smoothed shapes, or fragmented surfaces.
  }
  \label{fig:qualitative_comparison}
\end{figure}

Table~\ref{tab:toys4k_3col} and Figure~\ref{fig:qualitative_comparison} evaluate single-turn generation under both text- and image-conditioned settings on Toys4K. These two settings stress complementary capabilities: text-to-3D requires mapping abstract language to plausible geometry without dense spatial evidence, whereas image-to-3D rewards faithful reconstruction from pixel-aligned visual cues. EVA01 is designed to support both regimes within one mesh-native sequence model, rather than relying on separate task-specific pipelines.

\textbf{Bridging the Semantic Gap in Text-to-3D.}
In text-to-3D, EVA01 achieves \textbf{35.72} CLIP, \textbf{122.48} FD, \textbf{1.18} KD, and \textbf{70.4\%} preference, outperforming the strongest baseline TRELLIS (30.80 CLIP, 238.45 FD, 4.25 KD, 14.8\% preference) and the text-specialized 3DGen-R1 (29.35 CLIP, 263.72 FD, 6.85 KD, 7.9\% preference). This large margin indicates that dense-reconstruction backbones, although effective when visual evidence is available, remain inefficient at converting discrete language semantics into coherent 3D structure. EVA01 narrows this semantic-to-geometric gap by routing language-conditioned reasoning through the Understanding Expert ($E_{\text{und}}$), while the Generation Expert ($E_{\text{gen}}$) synthesizes sparse 3D latents through our staged alignment bridge and image-to-mesh warm-up curriculum. Compared with autoregressive mesh-token generation such as ShapeLLM-Omni (FD 310.55, KD 18.20), the flow-matching decoder further avoids severe quantization artifacts and yields smoother, more complete topology.

The right block of Figure~\ref{fig:qualitative_comparison} provides a more diagnostic view of this gap. The prompts contain unusual compositions and long attribute chains, such as a pipe-organ levitation train, a wingless supersonic vehicle, an anechoic wedge-cliff house, and a biomechanical clockwork heart. Baselines tend to satisfy only fragments of these descriptions: GVGEN often degenerates into sparse disconnected pieces, Michelangelo produces over-smoothed or incomplete gray geometry, and reconstruction-oriented methods such as 3DTopia-XL, ShapeLLM-Omni, and TRELLIS frequently recover local parts while missing the intended global object category or material organization. EVA01 is not merely sharper at the surface level; it more consistently preserves the requested object identity, assembles semantically related parts into a coherent whole, and produces paired texture/normal outputs that remain structurally aligned. This behavior supports the role of our bridge training: language semantics are first grounded in the MLLM representation space and then transferred to the sparse 3D latent space, rather than being learned as a weak text condition attached to a reconstruction model.

\textbf{High-Fidelity Image-to-3D.}
For image-to-3D, TRELLIS.2 remains the strongest specialist model, achieving \textbf{89.34} CLIP, \textbf{56.82} FD, \textbf{0.49} KD, and \textbf{41.7\%} preference. EVA01 ranks second with 87.28 CLIP, 61.74 FD, 0.63 KD, and 22.2\% preference, ahead of Hunyuan3D-2.1, Step1X-3D, Direct3D-S2, Hi3DGen, and the original TRELLIS in the overall metric profile. This ranking is expected: image-to-3D is primarily a dense reconstruction problem, and TRELLIS.2 is explicitly optimized to exploit pixel-aligned visual evidence from a single image. EVA01 instead uses the same MoT architecture and sparse latent interface for text generation, image generation, and downstream editing; its result therefore measures how much dense visual fidelity can be retained without abandoning a unified 3D-native MLLM formulation.

The left block of Figure~\ref{fig:qualitative_comparison} makes this trade-off visible. For character, robot, vehicle, and articulated object inputs, weak baselines either lose major structural parts, scatter geometry into disconnected fragments, or produce texture/normal predictions that no longer correspond to the same underlying shape. Michelangelo often recovers a coarse gray volume but weakens material and part separation; 3DTopia-XL and Hunyuan3D-2.1 can produce plausible local components but show instability in global assembly; ShapeLLM-Omni, Step1X-3D, and TRELLIS preserve more recognizable structure yet still introduce missing limbs, duplicated subparts, or inconsistent normal maps in challenging cases. EVA01 generally maintains the input object's category, pose, and large-scale topology while producing texture and normal renderings that remain aligned across views. The remaining gap to TRELLIS.2 reflects the current limit of our SigLIP2/DeepStack visual path rather than a failure of the MoT formulation; we analyze this limitation and discuss concrete DINOv3-inspired improvements in Section~\ref{sec:limitations}.

\subsection{Mesh Understanding}

% Requires: booktabs, xcolor with table option, colortbl, multirow, graphicx.

\begin{table}[!htbp]
\centering
\caption{
\textbf{Quantitative evaluation of mesh understanding on PointLLM-200.}
All models are evaluated on the PointLLM-200 Objaverse captioning benchmark with the same prompt:
\emph{``Caption this 3D model in detail.''}
We report traditional lexical overlap metrics, embedding-based semantic metrics, and two GPT-5.5 judge scores.
\textbf{GPT-ref} scores captions against the human reference text, while
\textbf{GPT-img} scores captions directly against multi-view RGB renders without using the reference caption.
All reported metrics are computed with the same scorer within each metric family, and higher is better.
}
\label{tab:pointllm200_captioning_results}
\setlength{\tabcolsep}{1.2pt}
\renewcommand{\arraystretch}{1.08}
\arrayrulecolor{eva_purple}
\resizebox{0.94\linewidth}{!}{%
\begin{tabular}{@{}l*{8}{c}@{}}
\toprule
\rowcolor{eva_purple}
\textcolor{white}{\textbf{Model}}
& \multicolumn{4}{c}{\textcolor{white}{\textbf{Traditional Metrics}}}
& \multicolumn{2}{c}{\textcolor{white}{\textbf{Semantic Metrics}}}
& \multicolumn{2}{c}{\textcolor{white}{\textbf{Judge Metrics}}} \\
\rowcolor{eva_purple}
\textcolor{white}{}
& \textcolor{white}{\textbf{B-1 $\uparrow$}}
& \textcolor{white}{\textbf{B-4 $\uparrow$}}
& \textcolor{white}{\textbf{R-L $\uparrow$}}
& \textcolor{white}{\shortstack[c]{\textbf{METEOR}\\$\uparrow$}}
& \textcolor{white}{\textbf{SBERT $\uparrow$}}
& \textcolor{white}{\textbf{SimCSE $\uparrow$}}
& \textcolor{white}{\textbf{GPT-ref $\uparrow$}}
& \textcolor{white}{\textbf{GPT-img $\uparrow$}} \\
\midrule
GT caption
  & -- & -- & -- & -- & -- & -- & -- & 56.050 \\
\midrule
PointLLM-7B
  & 7.905  & 0.669 & 10.290 & 13.536 & 47.894 & 48.812 & \underline{52.360} & 51.495 \\
PointLLM-13B
  & 7.873  & 0.649 & 10.519 & 13.620 & 47.539 & 48.602 & 51.735 & 49.745 \\
ShapeLLM-7B
  & 11.491 & 1.222 & 14.035 & 13.316 & 37.693 & 38.047 & 24.560 & 30.075 \\
ShapeLLM-13B
  & 10.542 & 1.050 & 12.954 & 14.234 & 39.935 & 40.728 & 33.925 & 35.870 \\
MiniGPT-3D
  & 5.482  & 0.468 & 7.680  & 10.531 & 28.708 & 23.722 & 17.360 & 7.560 \\
LLaMA-Mesh
  & 5.119  & 0.441 & 7.439  & 9.904  & 26.326 & 25.095 & 10.740 & 7.085 \\
GreenPLM
  & \underline{17.527} & \underline{2.303} & \underline{20.632} & \textbf{19.906}
  & 47.826 & 48.871 & 50.305 & 50.475 \\
ShapeLLM-Omni
  & 11.326 & 1.197 & 14.190 & 13.276 & 34.617 & 35.115 & 25.625 & 20.190 \\
\midrule
\rowcolor{eva_mint}
EVA01-Align
  & \textbf{23.592} & \textbf{3.555} & \textbf{28.002} & \underline{19.007}
  & \textbf{52.178} & \textbf{52.005} & 48.825 & \underline{56.770} \\
\rowcolor{eva_mint}
EVA01-Final
  & 6.334  & 0.585 & 9.323  & 13.566
  & \underline{50.554} & \underline{50.878} & \textbf{59.095} & \textbf{65.910} \\
\bottomrule
\end{tabular}%
}
\arrayrulecolor{black}
\end{table}

Table~\ref{tab:pointllm200_captioning_results} reveals two complementary aspects of mesh understanding: reference-aligned captioning and open-ended semantic coverage. The model after the first three curriculum stages, denoted \textbf{EVA01-Align}, achieves the strongest reference-aligned captioning performance among all evaluated model predictions. It obtains the best BLEU-1, BLEU-4, ROUGE-L, Sentence-BERT, and SimCSE scores, and ranks second on METEOR. This confirms that the first three stages of our curriculum successfully establish a strong mesh-language alignment pathway. In particular, the mesh understanding warm-up, visual-geometric initialization, and semantic modality alignment stages jointly map 3D geometry into the language space in a way that matches the official PointLLM-200 reference caption distribution more effectively than prior 3D understanding baselines.

However, the table also highlights an important limitation of standard captioning metrics for evaluating mesh understanding. BLEU, ROUGE-L, and METEOR primarily measure lexical overlap with a single reference caption. As a result, they favor predictions that closely match the wording, length, and style of the ground-truth caption. This is useful for measuring reference-style caption imitation, but it does not fully capture open-ended 3D understanding: the same mesh can be correctly described with different object names, attributes, part descriptions, or levels of detail. A model may mention valid geometric or visual details that are visible in the mesh but absent from the single reference caption, and such details can reduce $n$-gram precision despite being semantically correct. We therefore interpret these traditional metrics as measuring alignment to the PointLLM-200 caption style, rather than as complete measures of semantic mesh understanding.

This distinction explains the behavior of \textbf{EVA01-Final}. After the final instruction-oriented and high-quality finetuning stages, its lexical overlap scores decrease substantially, indicating that its output distribution shifts away from short, reference-like captions. At the same time, EVA01-Final retains the second-best Sentence-BERT and SimCSE scores among model predictions, and achieves the highest score under both GPT-ref and GPT-img. As an auxiliary sanity check, we also score the official reference caption itself with GPT-img. Both EVA01-Align and EVA01-Final score above this GT-caption row (56.77 and 65.91 vs. 56.05), where the judge compares each candidate caption directly against multi-view renders rather than against the reference sentence. This does not mean that the official reference captions are incorrect; rather, the PointLLM-200 references are brief human-annotated captions intended as single-reference evaluation targets, while GPT-img rewards detailed visual coverage of the rendered 3D object.

These results suggest that the later stages do not erase the mesh semantics learned by EVA01-Align. Instead, they repurpose this grounding for more instruction-following, fine-grained, and human-preferred descriptions. This behavior is consistent with prior findings in instruction tuning, where models trained to follow natural-language instructions become better aligned with user intent and human preference, even when their outputs diverge from narrow reference text patterns~\cite{ouyang2022training,chung2022scaling}. In our setting, the final model trades strict reference-caption overlap for broader semantic coverage, which is better captured by the GPT-based judges.

Among external baselines, GreenPLM~\cite{tang2025more} is the strongest reference-overlap competitor. It achieves the best baseline BLEU-1, BLEU-4, and ROUGE-L scores, and obtains the highest METEOR score overall. Under the GPT-based judges, however, PointLLM-7B is the strongest external baseline, while GreenPLM remains close. Nevertheless, EVA01-Align surpasses GreenPLM on most overlap and embedding-based metrics, while EVA01-Final exceeds all external baselines by a large margin under both reference-based and render-grounded GPT evaluation. The comparison with PointLLM also shows that simply scaling the language backbone is insufficient: PointLLM-7B and PointLLM-13B perform similarly across most metrics, suggesting that effective 3D-language alignment is more important than language-model size alone. Meanwhile, methods whose released inference interfaces are less matched to the PointLLM-200 point-cloud captioning protocol, including OBJ-text serialization in LLaMA-Mesh and the released 3D MLLM checkpoints of MiniGPT-3D and ShapeLLM variants, transfer less effectively under this unified prompt-only evaluation.

\begin{figure}[!htbp]
    \centering
    \includegraphics[width=0.92\linewidth]{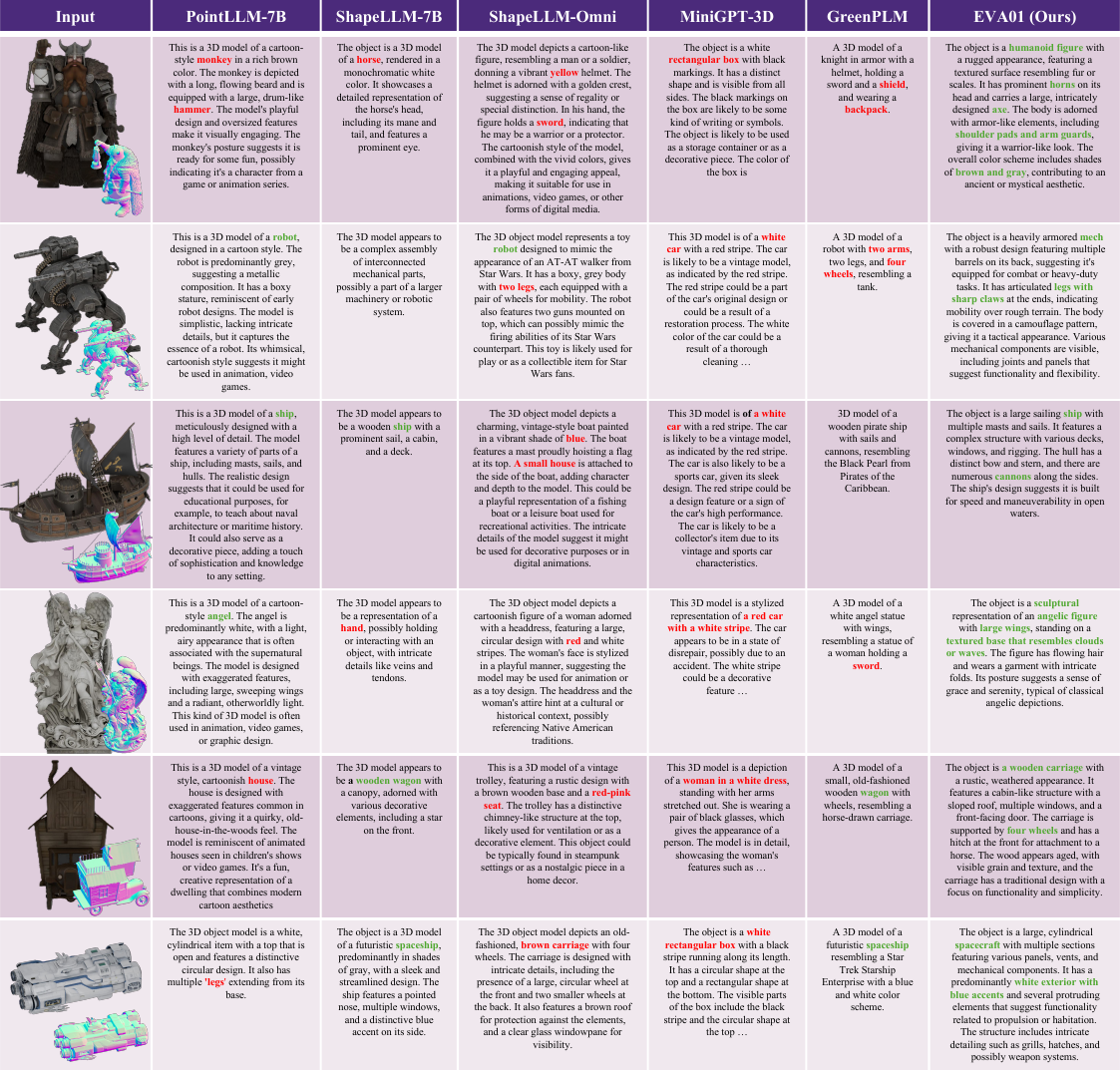}
    \caption{Qualitative comparison of mesh understanding results on PointLLM-200.}
    \label{fig:und_comparison}
\end{figure}

\begin{figure}[!htbp]
    \centering
    \includegraphics[width=0.86\linewidth,height=0.72\textheight,keepaspectratio]{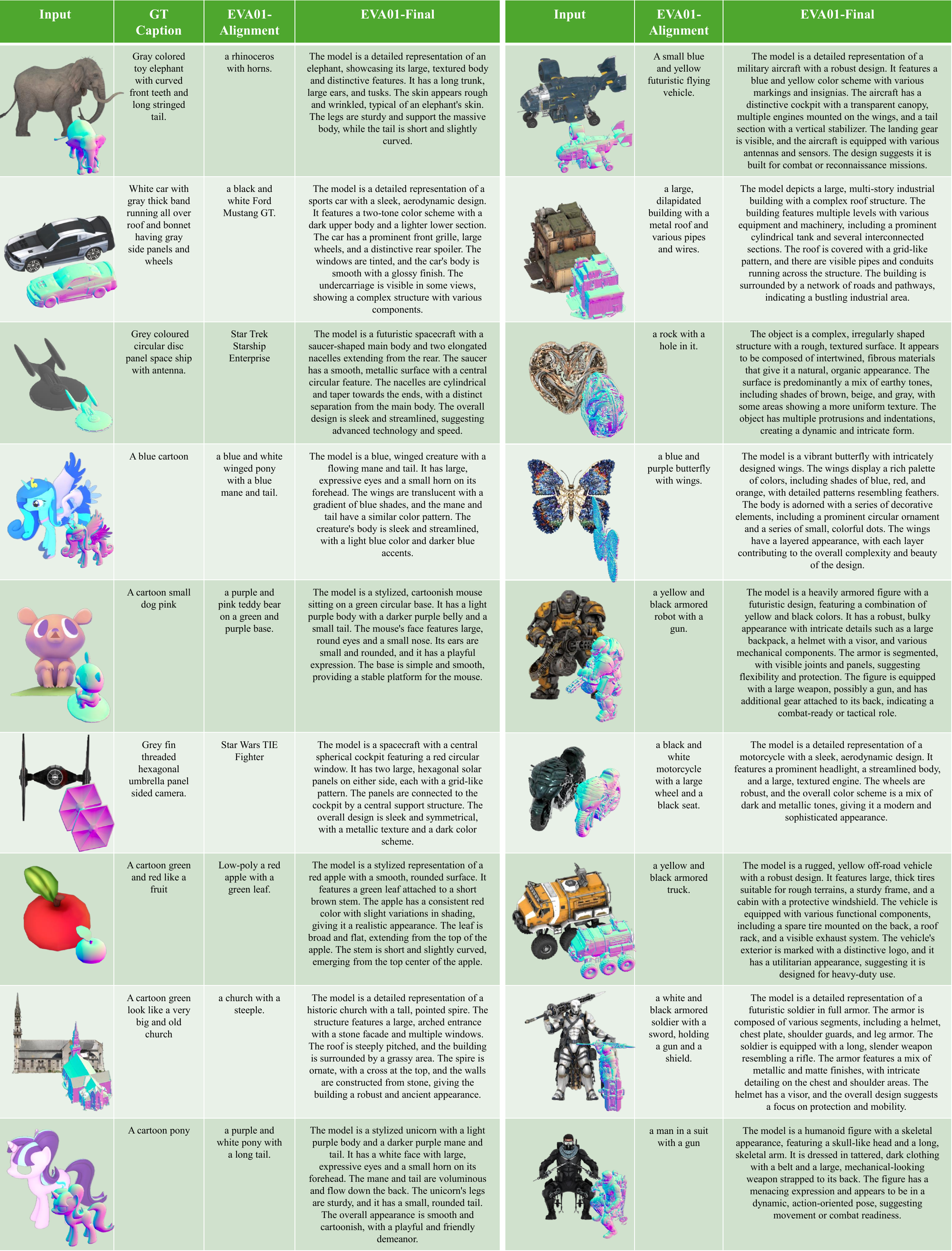}
    \caption{\textbf{Additional Mesh Understanding Examples.}
    Left: randomly sampled PointLLM-200 examples with the rendered input, official reference caption, EVA01-Align caption, and EVA01-Final caption.
    Right: captions for generated 3D models without ground-truth annotations.
    EVA01-Final provides richer part-, material-, color-, and structure-level descriptions than the alignment-stage model.}
    \label{fig:und_full}
\end{figure}

Overall, these results support the intended role of our curriculum. The first three stages produce a strong reference-aligned mesh captioner, demonstrating successful alignment between geometric input and language, but the qualitative examples in Figures~\ref{fig:und_comparison} and~\ref{fig:und_full} show that EVA01-Align often stays at the level of object identity and coarse attributes. The final two stages adapt this understanding pathway to the broader unified setting, where the model must follow instructions and support generation/editing behavior rather than merely imitate short captions. As shown in Figure~\ref{fig:und_full}, EVA01-Final produces richer descriptions of parts, materials, colors, and local structure, explaining its stronger GPT-img score. This gain comes with the usual instruction-tuning trade-off: the model may occasionally infer details not explicitly visible in the rendered input. The final model is therefore not optimized for maximum lexical overlap with a single reference, but for fine-grained, visually grounded semantic coverage.

\subsection{Multi-Turn Context-Aware Editing}

\par A defining capability of EVA01 is native, multi-turn 3D editing. Unlike cascaded pipelines that process each edit as an independent reconstruction problem, EVA01 models editing as a \textbf{stateful conditional sequence generation task}: the next mesh state is predicted from the current instruction and the accumulated text--image--mesh history. This formulation is essential for long-horizon editing, where the model must both apply a localized change and preserve the identity, topology, material layout, and previous edits of the same object.

\begin{figure}[!htbp]
  \centering
  \includegraphics[width=0.82\linewidth]{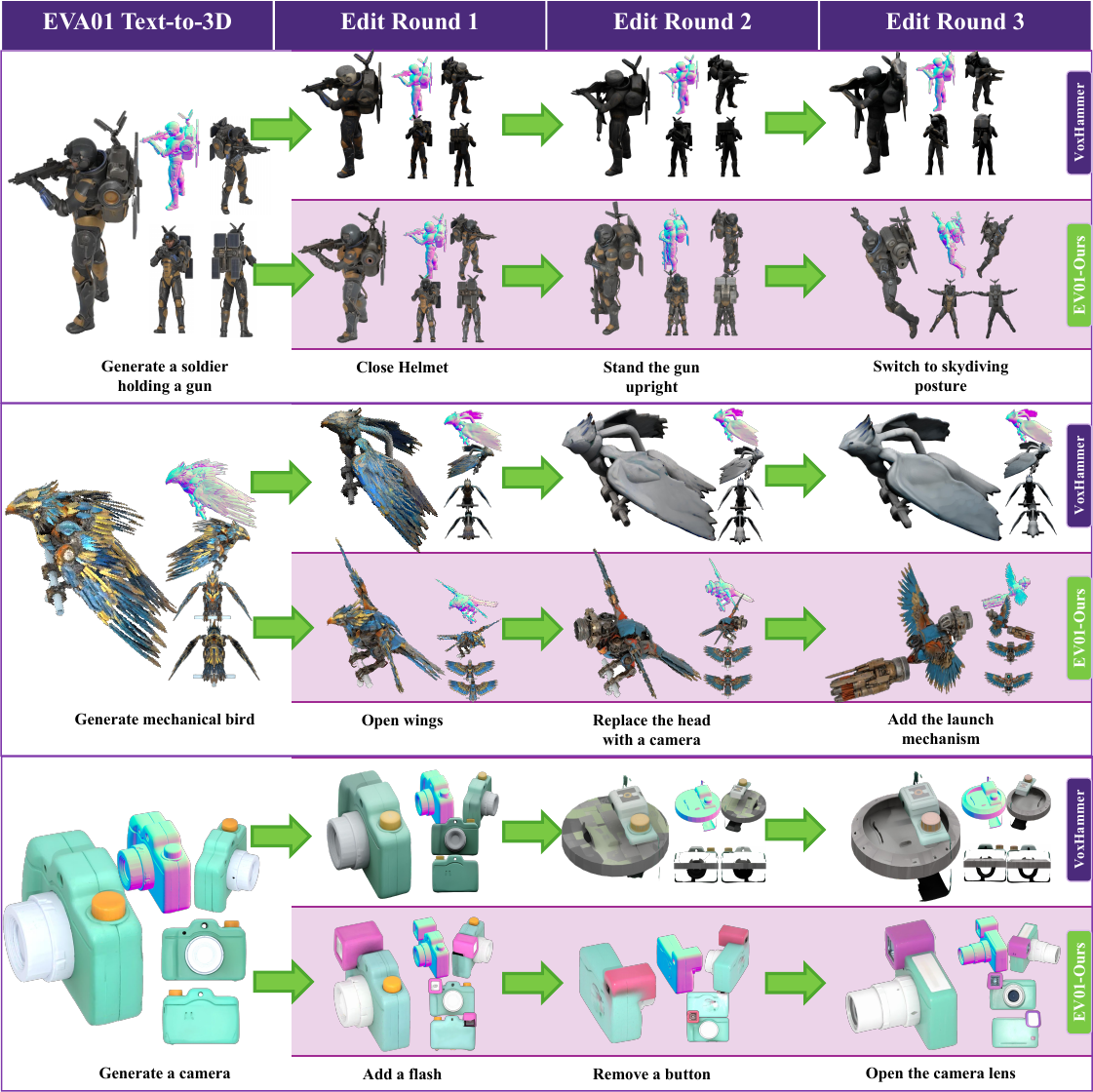}
  \caption{
  \textbf{Versatile Instruction-Driven 3D Editing Gallery.}
  Starting from EVA01 text-to-3D generations, we compare three rounds of sequential edits against VoxHammer.
  Across a camera, a mechanical bird, and a soldier, EVA01 accumulates instructions---adding or removing parts, changing object state, replacing components, and modifying pose---while preserving object identity and geometry history.
  Additional long-horizon trajectories are shown in Figure~\ref{fig:additional_demos}.
  }
  \label{fig:versatile_editing_gallery}
\end{figure}

\textbf{Stateful Editing without Explicit Masks.}
EVA01 enforces the block attention mask in Table~\ref{tab:attn_mask}, allowing the current noisy mesh tokens to attend to clean historical mesh states while preventing future leakage. Together with 3D Interleaved MRoPE, this gives each sparse latent token both a temporal role in the interaction and a spatial coordinate in the 3D grid. As a result, the model learns a differential geometric update conditioned on history, rather than regenerating the object from scratch. This is the main distinction from mask-conditioned baselines: Instant3DiT~\cite{barda2025instant3dit} edits through multiview inpainting followed by reconstruction, TRELLIS~\cite{xiang2024structured} uses a structured reconstruction backbone for mask-guided editing, and VoxHammer~\cite{li2025voxhammer} performs native 3D editing with explicit edit masks. EVA01 receives only the instruction and historical context at inference time.

\begin{table}[tb]
\centering
\caption{\textbf{Multi-turn editing evaluation.} CD and PSNR evaluate unedited-region consistency; CLIP, FD$_{\text{DINOv2}}$, and Pref\% evaluate overall editing quality.}
\label{tab:multi_turn_editing_eval}
\small
\setlength{\tabcolsep}{2.5pt}
\renewcommand{\arraystretch}{1.08}
\arrayrulecolor{eva_purple}
\begin{tabular}{@{}>{\raggedright\arraybackslash}p{0.29\linewidth}*{5}{>{\centering\arraybackslash}p{0.115\linewidth}}@{}}
\toprule
\rowcolor{eva_purple}
\textcolor{white}{\textbf{Method}}
& \multicolumn{2}{c}{\textcolor{white}{\textbf{Consistency}}}
& \multicolumn{3}{c}{\textcolor{white}{\textbf{Quality}}} \\
\rowcolor{eva_purple}
\textcolor{white}{}
& \textcolor{white}{\textbf{CD $\downarrow$}}
& \textcolor{white}{\textbf{PSNR $\uparrow$}}
& \textcolor{white}{\textbf{CLIP $\uparrow$}}
& \textcolor{white}{\textbf{FD $\downarrow$}}
& \textcolor{white}{\textbf{Pref.\% $\uparrow$}} \\
\midrule
Instant3DiT\textsuperscript{*}~\cite{barda2025instant3dit} & 0.034 & 21.4 & 24.68 & 265.34 & 0.00\% \\
TRELLIS\textsuperscript{*}~\cite{xiang2024structured} & 0.058 & 18.2 & 26.12 & 241.78 & 2.50\% \\
VoxHammer\textsuperscript{*}~\cite{li2025voxhammer} & 0.015 & 32.1 & \underline{30.45} & \underline{178.62} & \underline{3.75\%} \\
\rowcolor{eva_mint}
\textbf{Ours} & 0.018 & 29.3 & \textbf{70.18} & \textbf{89.37} & \textbf{93.75\%} \\
\bottomrule
\end{tabular}
\vspace{2pt}
\parbox{\linewidth}{\footnotesize \textsuperscript{*}Requires an explicit 3D edit mask at inference time; EVA01 does not use masks.}
\arrayrulecolor{black}
\end{table}

Table~\ref{tab:multi_turn_editing_eval} shows a clear separation between preservation-oriented consistency and instruction-following edit quality. VoxHammer obtains the best CD and PSNR (0.015 and 32.1), which is expected because it is given an explicit 3D edit mask and is designed to preserve unedited regions. EVA01 remains close in consistency (0.018 CD and 29.3 PSNR) despite using no mask, indicating that historical mesh tokens provide a strong implicit preservation signal. More importantly, EVA01 dominates the quality-oriented metrics: it improves CLIP from VoxHammer's 30.45 to \textbf{70.18}, reduces FD$_{\text{DINOv2}}$ from 178.62 to \textbf{89.37}, and receives \textbf{93.75\%} user preference, compared with 3.75\% for VoxHammer and 2.50\% for TRELLIS. This gap indicates that mask-based preservation alone is insufficient for multi-turn editing; the model must also understand the evolving object state and synthesize the requested new geometry in context.

\begin{figure}[!htbp]
  \centering
  \includegraphics[width=\linewidth,height=0.72\textheight,keepaspectratio]{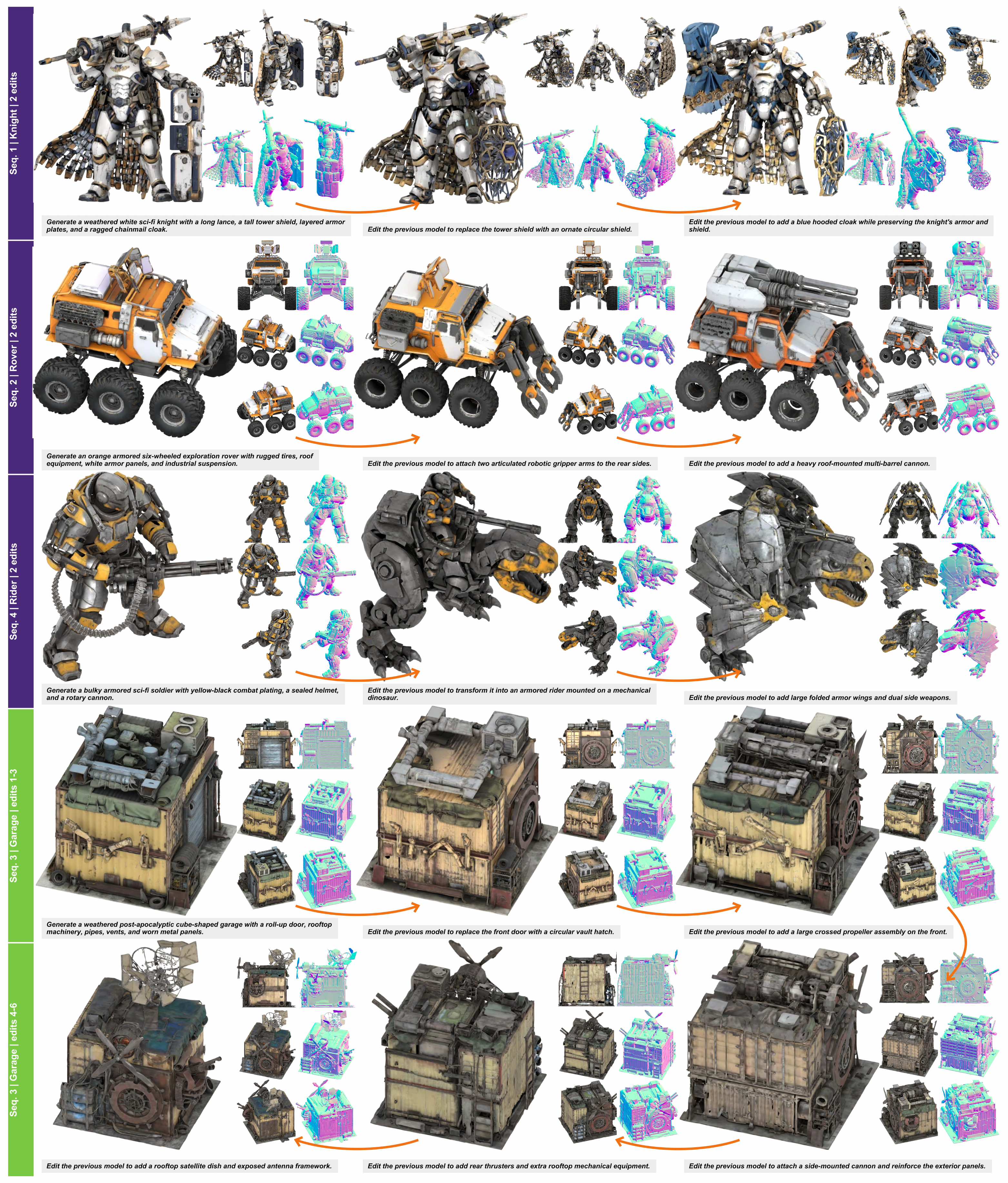}
  \caption{\textbf{Additional Generation and Editing Results.}
  Each cell shows a trajectory state with a main render, auxiliary RGB/normal views, and a concise generation or editing prompt.}
  \label{fig:additional_demos}
\end{figure}

Figures~\ref{fig:versatile_editing_gallery} and~\ref{fig:additional_demos} illustrate the same conclusion qualitatively. In the camera, bird, and soldier examples, EVA01 accumulates edits across three rounds while preserving the source identity: added parts remain attached in later turns, removed components do not reappear, and pose or articulation changes are applied without destroying previously edited structure. The additional long-horizon trajectories further stress harder edits, including replacing a knight's shield before adding a cloak, attaching robotic arms and a roof cannon to a rover, transforming a soldier into a rider on a mechanical dinosaur, and repeatedly modifying a garage with a vault hatch, propeller, satellite dish, thrusters, and reinforced exterior panels. These examples require constructive geometry, part replacement, articulation, accessory insertion, and material preservation over two to five turns. EVA01's ability to keep RGB and normal renderings aligned across these trajectories suggests that it maintains a coherent latent state rather than merely producing visually plausible single-step outputs.

\subsection{Subjective Evaluation}

\noindent\textbf{Study Protocol.}
We conduct user studies across all three evaluation axes: single-turn text-to-3D generation, single-turn image-to-3D generation, and multi-turn editing.
A total of 23 participants are recruited; each completes at least 25 paired or multi-way comparisons per category.
For each comparison, participants are shown the input condition (text prompt, reference image, or editing instruction with interaction history) alongside anonymized outputs from competing methods, presented in randomized order.
Participants are instructed to select the output that best satisfies the given condition, judging geometric fidelity, semantic alignment, and---for editing tasks---identity preservation across turns.
All selections are aggregated per category, and preference scores are reported as the percentage of total selections each method receives within each category.

\noindent\textbf{Results.}
Table~\ref{tab:toys4k_3col} reports preference results for single-turn generation.
In text-to-3D, EVA01 receives~\textbf{70.4\%} of all selections, a margin of nearly 5$\times$ over TRELLIS~(14.8\%) and over 8$\times$ over the text-specialized 3DGen-R1~(7.9\%).
In image-to-3D, TRELLIS.2 leads with 41.7\%; EVA01 places second at~\textbf{22.2\%}, ahead of Hunyuan3D-2.1~(10.2\%), Step1X-3D~(7.9\%), and other supported baselines.
Notably, EVA01 is the only method to rank among the top two in both modalities within a single unified model.

Table~\ref{tab:multi_turn_editing_eval} reports preference for multi-turn editing.
EVA01 receives~\textbf{93.75\%} of selections, compared to 3.75\% for VoxHammer and 2.50\% for TRELLIS, despite both baselines having access to explicit 3D edit masks while EVA01 operates mask-free.
This large preference gap reflects the advantage of stateful editing: EVA01 conditions each edit on the full interaction history, preserving geometric identity across turns, whereas mask-based baselines treat each edit as an independent reconstruction.
Participants consistently preferred EVA01's outputs for maintaining object identity while applying the requested structural changes across sequential editing instructions.

\subsection{Representation Analysis and Visualization}

\begin{table}[!htbp]
\centering
\small
\setlength{\tabcolsep}{5pt}
\renewcommand{\arraystretch}{1.12}
\caption{\textbf{Feature probing across visual and image-generation encoders.} Higher values are better for NAVI R@5cm, NYUv2 $\delta_1$, and Objaverse retrieval R@5; lower values are better for NYUv2 RMSE. Bold and underline indicate the best and second-best entries within each category block. \textcolor{eva_purple}{Lavender} marks semantic/understanding-token paths; \textcolor{eva_green}{mint} marks dense visual or generation-latent paths. ``--'' denotes metrics not applicable to raw visual/latent-only features in the retrieval protocol.}
\label{tab:feature_analysis_probe3d}
\begin{tabular}{l>{\centering\arraybackslash}p{2.5cm}rrrr}
\toprule
\rowcolor{eva_purple}
\textcolor{white}{\textbf{Model}} & \textcolor{white}{\textbf{Feature path}} & \textcolor{white}{\textbf{NAVI R@5cm$\uparrow$}} & \textcolor{white}{\textbf{NYUv2 $\delta_1{\uparrow}$}} & \textcolor{white}{\textbf{NYUv2 RMSE$\downarrow$}} & \textcolor{white}{\textbf{Retrieval R@5$\uparrow$}} \\
\midrule
\rowcolor{eva_purple}
\multicolumn{6}{l}{\textcolor{white}{\textbf{Visual baselines}}} \\
CLIP ViT-L/14-336 & final & 47.96 & 0.3859 & 1.1866 & \textbf{43.50} \\
SigLIP2 SO400M & final & 78.30 & 0.6659 & 0.6567 & \underline{23.00} \\
DINOv2 ViT-L/14 & final & \underline{91.85} & \underline{0.8772} & \underline{0.3341} & -- \\
DINOv3 ViT-L/16 & final & \textbf{92.74} & \textbf{0.8935} & \textbf{0.3126} & -- \\
\rowcolor{eva_purple}
\multicolumn{6}{l}{\textcolor{white}{\textbf{Qwen3-VL DeepStack / LLM}}} \\
\rowcolor{eva_mint}
Qwen3-VL-2B & VT ds5 & 66.04 & 0.4797 & 0.9802 & -- \\
\rowcolor{eva_mint}
Qwen3-VL-2B & VT ds17 & \underline{78.37} & \underline{0.7510} & 0.5612 & -- \\
\rowcolor{eva_mint}
Qwen3-VL-2B & VT final & 69.54 & 0.6897 & 0.6308 & -- \\
\rowcolor{eva_lavender}
Qwen3-VL-2B & LLM l07 & 75.23 & 0.7439 & \underline{0.5460} & 0.50 \\
\rowcolor{eva_lavender}
Qwen3-VL-2B & LLM l14 & 73.88 & \textbf{0.7826} & \textbf{0.5246} & 0.50 \\
\rowcolor{eva_lavender}
Qwen3-VL-2B & LLM l28 & 71.77 & 0.7396 & 0.5868 & \textbf{15.80} \\
\rowcolor{eva_mint}
Qwen3-VL-4B & VT ds17 & 77.79 & 0.6842 & 0.6371 & -- \\
\rowcolor{eva_lavender}
Qwen3-VL-4B & LLM l14 & 75.40 & 0.7458 & 0.5569 & 0.50 \\
\rowcolor{eva_mint}
Qwen3-VL-8B & VT ds16 & \textbf{80.25} & 0.7155 & 0.5975 & -- \\
\rowcolor{eva_lavender}
Qwen3-VL-8B & LLM l07 & 76.68 & 0.7474 & 0.5542 & 0.50 \\
\rowcolor{eva_purple}
\multicolumn{6}{l}{\textcolor{white}{\textbf{Image-generation MLLMs}}} \\
\rowcolor{eva_lavender}
Bagel-7B-MoT & ViT final & \textbf{76.45} & \textbf{0.6320} & \textbf{0.7009} & -- \\
\rowcolor{eva_mint}
Bagel-7B-MoT & VAE clean & 53.56 & 0.3929 & 1.1725 & -- \\
\rowcolor{eva_mint}
Bagel-7B-MoT & VAE -> LLM l21 & 61.11 & \underline{0.5923} & \underline{0.7249} & 0.60 \\
\rowcolor{eva_lavender}
Qwen-Image-Edit & Qwen2.5-VL final & 62.21 & 0.5901 & 0.7620 & \textbf{17.50} \\
\rowcolor{eva_mint}
Qwen-Image-Edit & MMDiT 20/30/40 & \underline{68.51} & 0.5509 & 0.8318 & 0.60 \\
\rowcolor{eva_mint}
Qwen-Image-Edit & Ref VAE latent & 55.04 & 0.3896 & 1.1877 & -- \\
\bottomrule
\end{tabular}
\end{table}

\begin{figure}[!htbp]
  \centering
  \includegraphics[width=\linewidth]{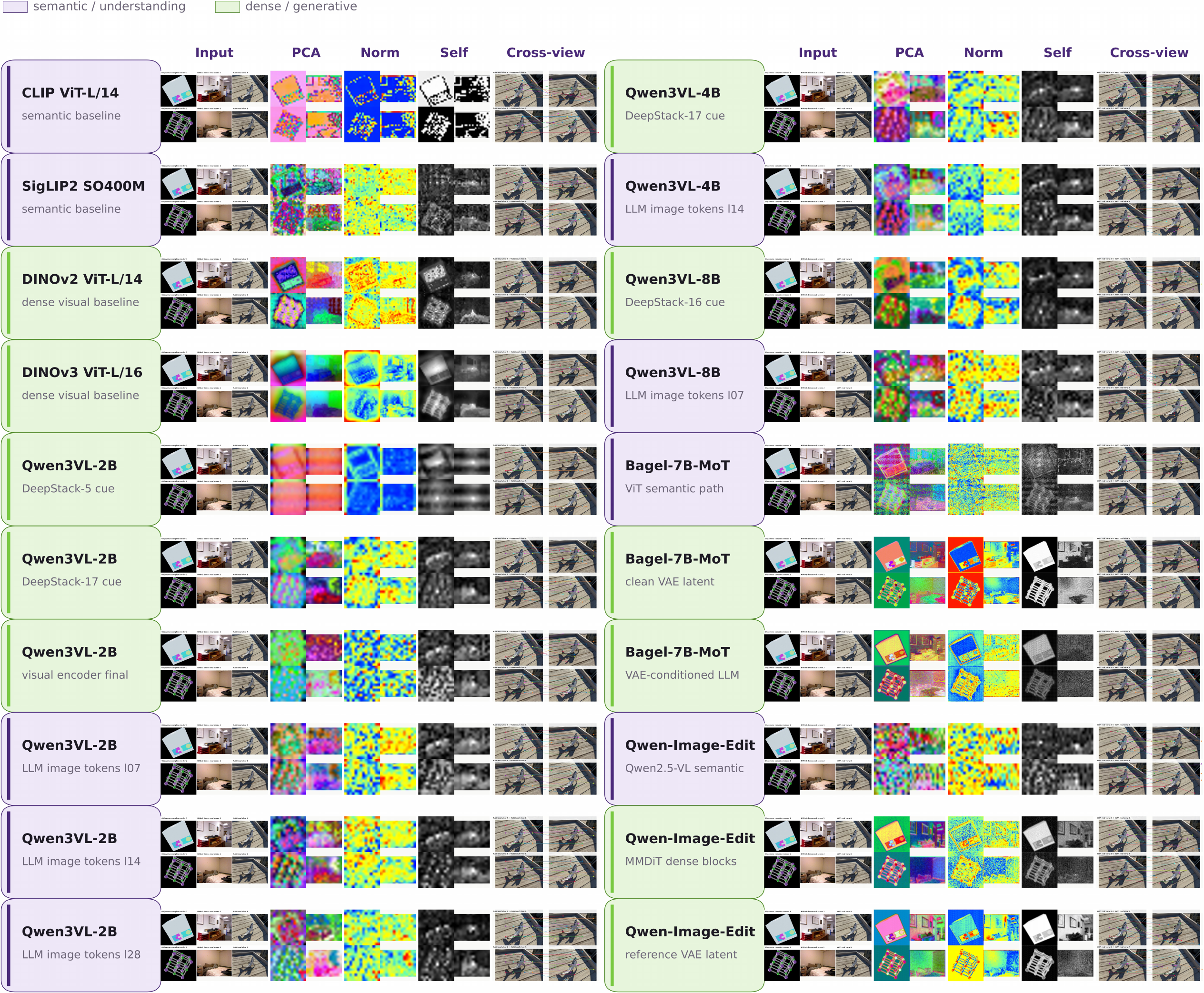}
  \caption{
  \textbf{Representation visualization across visual and image-generation encoders.}
  We visualize the same feature paths probed in Table~\ref{tab:feature_analysis_probe3d} using input views, PCA projections, normalized activation maps, self-similarity, and cross-view correspondence overlays.
  \textcolor{eva_purple}{Lavender} rows denote semantic or understanding-token paths, while \textcolor{eva_green}{mint} rows denote dense visual or generation-side latent paths.
  The visualization reveals that global semantic alignment, dense spatial correspondence, and generation-side appearance latents form distinct representation regimes rather than a single universally optimal feature space.
  }
  \label{fig:feature_analysis_visualization}
\end{figure}

Following recent protocols for probing 3D awareness in foundation models~\cite{el2024probing,huang2025much}, Table~\ref{tab:feature_analysis_probe3d} and Figure~\ref{fig:feature_analysis_visualization} evaluate which feature paths can provide visual, semantic, or generative conditioning for a 3D-native MLLM. NAVI R@5cm and the NYUv2 depth probes measure dense spatial and geometric information: NAVI tests whether cross-view correspondences fall within a 5cm 3D threshold, while NYUv2 $\delta_1$ and RMSE measure how linearly scene layout, depth ordering, and geometric boundaries can be read from frozen features. Objaverse retrieval R@5 follows a different protocol: image and caption features are mean-pooled, normalized, and matched by top-5 paired-caption retrieval. For contrastive models such as CLIP and SigLIP2, this directly probes the native image-text embedding space; for Qwen3-VL, Bagel, and Qwen-Image-Edit, it is a narrower test of whether hidden states can be pooled into retrieval embeddings, rather than a general measure of semantic understanding.

The visual baselines separate global semantic alignment from dense geometric structure. CLIP obtains the highest retrieval score (43.50), but weak NAVI and NYUv2 results (47.96, 0.3859, and 1.1866). Figure~\ref{fig:feature_analysis_visualization} explains this gap: CLIP's PCA maps capture coarse object-level regions and its self-similarity maps form high-contrast semantic blobs, whereas its norm maps are sparse and edge-biased, and its cross-view matches are less consistently anchored to local surface geometry. Thus, global image-text contrastive alignment does not by itself provide a stable correspondence field. In contrast, DINO-style dense features define the geometric upper envelope. DINOv3 ViT-L/16 achieves the strongest NAVI, $\delta_1$, and RMSE scores (92.74, 0.8935, and 0.3126); its PCA and norm maps are spatially smooth, object- and part-aligned, and preserve coherent room layout and object boundaries across Objaverse, NYUv2, and NAVI examples, with more geometrically plausible cross-view matches. DINOv2 shows the same dense-visual behavior with slightly weaker boundary and layout separation. SigLIP2 lies between the two regimes: although optimized as an image-text model, its maps collapse less than CLIP's and retain more local structure, consistent with its stronger NAVI and NYUv2 scores despite lower retrieval.

Qwen3-VL shows that 3D-relevant information is distributed across paths and layers rather than concentrated in the final hidden state. Along the visual-token path, middle-to-late DeepStack features preserve stronger local geometry: Qwen3-VL-8B VT ds16 gives the best NAVI score in the Qwen3-VL block (80.25), and Qwen3-VL-2B VT ds17 is close (78.37), both clearly above early VT ds5 and the visual-encoder final state. The visualization shows the same pattern: DeepStack PCA maps retain object and part regions, and cross-view matches remain tied to corresponding local surfaces, whereas early VT features are smoother and less discriminative, and the final visual state is more mixed. LLM image-token layers shift the representation toward scene-level reasoning. Qwen3-VL-2B LLM l14 gives the strongest NYUv2 result in the Qwen3-VL group (0.7826 $\delta_1$, 0.5246 RMSE), and its NYUv2 PCA and self-similarity panels emphasize broader room-layout regions rather than fine object parts. Later LLM layers move further toward global semantic compatibility: Qwen3-VL-2B LLM l28 has the highest retrieval score in this group (15.80), while its dense probing scores weaken. These trends favor layer- and path-aware routing over simply using the last hidden state or scaling the model.

Image-generation MLLMs exhibit a complementary limitation. In Bagel, the ViT semantic path is the strongest feature for dense probing, while the clean VAE latent is much weaker. The figure makes this distinction explicit: clean VAE latents preserve appearance, reconstruction layout, and high-frequency edges in the PCA, norm, and self-similarity maps, especially on synthetic Objaverse objects, but this reconstruction-oriented signal does not translate into robust NAVI or NYUv2 geometry. Passing VAE latents through the LLM partially restores higher-level structure and improves probing scores, yet remains below the best visual and Qwen3-VL paths. Qwen-Image-Edit shows the same division of labor. Its Qwen2.5-VL final state is better suited to semantic compatibility and retrieval, whereas MMDiT blocks expose denser local cues: their PCA and self-similarity maps preserve object silhouettes, edges, and scene layout more clearly than the purely semantic path. The reference VAE latent again behaves as an appearance/reconstruction condition, with crisp contours but weak transferable geometric abstraction. Generation-side latents are therefore useful for synthesis and appearance anchoring, but should not be treated as complete 3D geometry representations.

These results motivate EVA01's separation of semantic reasoning from geometric generation. A 3D-native system cannot reduce visual conditioning to a single frozen encoder, final MLLM hidden state, or VAE latent. Table~\ref{tab:feature_analysis_probe3d} shows that semantic retrieval, cross-view correspondence, and depth probing peak in different feature families; Figure~\ref{fig:feature_analysis_visualization} shows the same split through coarse semantic blobs, dense object-aligned PCA fields, layout-sensitive self-similarity maps, and cross-view match stability. EVA01 therefore keeps the Understanding Expert as a stable semantic anchor while allowing the Generation Expert to access both semantic tokens and dense visual-geometric cues through shared global attention. The CLIP--DINO split, Qwen3-VL's DeepStack/LLM specialization, and the Bagel/Qwen-Image-Edit separation among VAE, transformer, and semantic paths all support the same design principle: semantic reasoning and geometric generation should be decoupled, yet remain communicative within a shared sequence space.

\subsection{Ablation Studies and Critical Insights}
\label{subsec:ablation}
\begin{figure}[!htbp]
  \centering
  \begin{minipage}[t]{0.5\linewidth}
    \vspace{0pt}
    \centering
    \includegraphics[width=\linewidth]{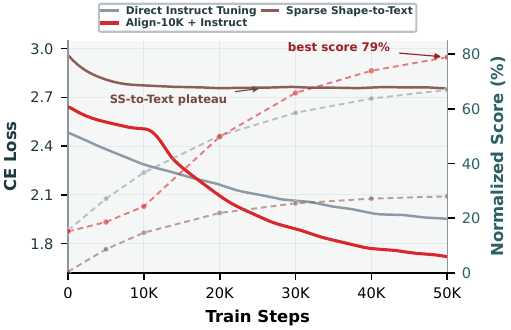}
  \end{minipage}%
  \begin{minipage}[t]{0.5\linewidth}
    \vspace{0pt}
    \centering
    \includegraphics[width=\linewidth]{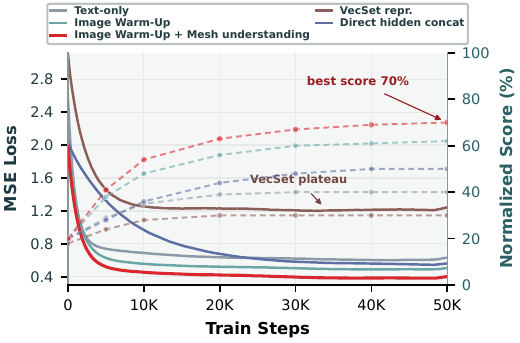}
  \end{minipage}
  \caption{\textbf{Training Dynamics and Loss Curves.}
  Left: mesh-understanding ablations comparing direct instruction tuning, a 10K alignment warm-up followed by instruction tuning, and Sparse Shape-to-Text, which uses generation-side sparse VAE latents for captioning. Solid curves report CE loss, and dashed marker curves report normalized captioning score. Right: Sparse Shape generation ablations comparing text-only training, image warm-up, image warm-up with mesh understanding, VecSet representation, and multi-layer hidden-feature concatenation for cross-attention. Solid curves report MSE loss, and dashed marker curves report normalized generation score relative to the best text-to-3D checkpoint.
  }
  \label{fig:loss_curves}
\end{figure}

Figure~\ref{fig:loss_curves} summarizes the ablations that shaped EVA01's final curriculum and architecture. The central question is not only whether a mesh representation can be decoded into geometry, but whether it can be aligned with the pre-trained MLLM semantic space, optimized inside a unified sequence, and reused for context-aware generation without destroying the language prior.

The mesh-understanding curves reveal that alignment must precede instruction tuning. Directly finetuning on mesh captions lowers CE loss rapidly in the first few thousand steps, but the curve soon saturates and reaches a weaker normalized captioning score. In contrast, the 10K alignment warm-up descends more slowly at the beginning, yet it creates a cleaner semantic bridge between Point-BERT mesh features and the Qwen3-VL token space. Once instruction tuning starts, this aligned model overtakes the direct-tuning baseline and converges to both lower CE loss and higher captioning quality. The Sparse Shape-to-Text variant exposes a complementary failure mode: directly using generation-side sparse VAE latents as the mesh-understanding input, even when the generation pathway is treated as an encoder and fully tuned, does not yield a stable captioning model. Its loss plateaus around a high CE regime and the generated descriptions remain semantically unreliable. This confirms that sparse generation latents are effective reconstruction variables, but they are not by themselves language-readable semantic tokens.

The generation curves show why the curriculum begins from image-conditioned generation before text-to-3D alignment. Training only from text reduces MSE, but it converges more slowly and reaches a lower score. Image warm-up provides a stronger bridge because dense visual features already live closer to the MLLM's semantic manifold and carry local geometric evidence through the DeepStack pathway. Adding mesh-understanding samples further improves both convergence and the final score, indicating that the captioning objective is not merely an auxiliary regularizer: it sharpens the spatial grounding of $E_{\text{und}}$, and shared global attention allows $E_{\text{gen}}$ to query these better-aligned semantic features during generation. This bidirectional coupling is also more effective than directly feeding multi-layer hidden features through concatenation-based cross-attention, which supplies additional conditioning but lacks token-level mutual visibility inside the unified sequence. As a result, it improves over text-only training but remains below the MoT formulation in both efficiency and upper bound, and it does not naturally support context-aware editing over interleaved mesh histories.

The same figure also explains why EVA01 adopts a structured sparse grid rather than a permutation-invariant VecSet representation~\cite{zhang2023shapetovecset}, including the variant used in Hunyuan3D-2.1~\cite{hunyuan3d2025hunyuan3d}. Under our unified sequence setting, VecSet fails to produce usable geometry: the loss quickly reaches a plateau and the normalized score remains far below grid-based sparse latents. Although VecSet is compact, its latent tokens do not carry an intrinsic position anchor after being flattened into the MLLM sequence. Attention therefore observes a weak global token identity, but cannot reliably distinguish where each latent token is located in 3D space. In contrast, our sparse voxel tokens bind each feature to an explicit coordinate $\boldsymbol{p}_i$, making 3D Interleaved MRoPE meaningful and allowing attention to model local and global spatial relations. These results support three practical design rules for 3D-native MLLMs: modality dropout is needed to prevent $E_{\text{gen}}$ from ignoring weak text conditions during Stage~3; Mesh-to-Text supervision provides a necessary bidirectional alignment signal rather than a secondary task; and high-quality finetuning in Stage~5 sets the final fidelity ceiling after the representation and alignment problems have been stabilized.

\begin{figure}[!htbp]
    \centering
    \includegraphics[width=1\linewidth]{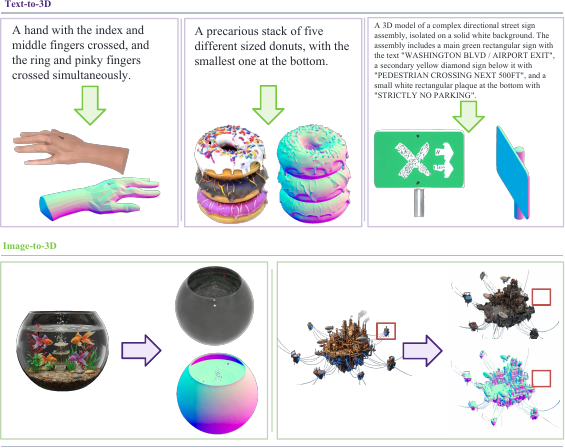}
    \caption{\textbf{Failure Cases of EVA01.}
    In text-to-3D generation, EVA01 still has limited generalization to out-of-distribution compositions and remains imperfect at spatial reasoning, exact counting, and producing legible text or symbol layouts on 3D surfaces.
    In image-to-3D cases, failure is often caused by insufficient dense pixel evidence in the input view: when thin structures, occluded parts, or small distant components are under-resolved, the model may recover the dominant object while losing local details or producing incomplete geometry.}
    \label{fig:badcase}
\end{figure}

\section{Limitations, Discussion \& Future Work}
\label{sec:limitations}

EVA01 is bounded by the training budget of this study: both experts operate at the 2B scale with $512^3$ sparse-voxel resolution. This setting suffices to verify the central design hypothesis---that semantic reasoning and geometric generation should be decoupled yet connected through shared global attention---but does not exhaust the scaling behavior of 3D-native MLLMs. Scaling each expert toward 4B--8B and increasing resolution to $1024^3$ is a natural next step, where larger context and finer spatial discretization are expected to improve both long-context reasoning and high-frequency geometric recovery.

Current automatic 3D generation metrics remain partial proxies for perceptual and geometric quality. Metrics built on aligned 3D representation spaces, such as ULIP, ULIP-2, and Uni3D~\cite{xue2023ulip,xue2024ulip,zhou2024uni3d}, measure coarse semantic agreement between language, images, and 3D shapes, but operate on sampled point clouds or global shape embeddings. They under-resolve surface-level properties critical for high-fidelity mesh generation: local topology, thin structures, sharp creases, watertightness, material-boundary consistency, and whether an edit preserves the identity of unmodified regions. This creates a mismatch between improving generators and comparatively coarse evaluators. Developing benchmarks for fine-grained mesh quality, semantic faithfulness, and context-aware editing consistency remains an important direction.

A third limitation is the representational asymmetry between mesh understanding and mesh generation. Even architectures that decouple visual encoders---Janus and Bagel~\cite{wu2024janus,deng2025bagel}---consume RGB images on both sides. In EVA01, the understanding branch relies on point-cloud features, while the generation branch operates on structured sparse-voxel latents. This split is pragmatic: point-cloud encoders provide a stable entry point for mesh captioning, whereas sparse voxels are better matched to high-resolution geometry synthesis and 3D positional attention. A more native interface would replace this split with a shared 3D latent substrate. Inspired by unified visual encoder families such as OpenVision and OpenVision~3~\cite{li2025openvision,zhang2026openvision}, future work should explore sparse-voxel-native encoder families in which understanding and generation share a common 3D representation served by specialized encoders over the same latent backbone.

The image-conditioned setting exposes another boundary, with representative failure modes shown in Figure~\ref{fig:badcase}. DeepStack injects lower-level SigLIP2 features into the MLLM backbone, but our representation analysis shows these features are weaker dense geometric carriers than DINO-style self-supervised visual features. Consequently, EVA01 is competitive in image-to-3D while trailing the strongest image-specialized reconstruction pipeline. This does not reflect a failure of the MoT formulation; rather, a 3D-native MLLM requires sufficiently local, patch-aligned visual evidence for the Generation Expert to reconstruct fine geometry from a single image. Replacing the current SigLIP2 visual path with encoders that jointly preserve text alignment and dense patch structure---such as TIPSv2~\cite{cao2026tipsv2}---or adopting a multi-tower visual aggregation design in the spirit of Cambrian-1's Spatial Vision Aggregator~\cite{tong2024cambrian} could strengthen the pixel-dense conditioning available to $E_{\text{gen}}$ without sacrificing semantic grounding.

Taken together, these limitations delineate the scope of this work. EVA01 does not claim that a 2B, $512^3$ instance saturates all 3D generation regimes, nor that existing metrics fully capture mesh quality. Its contribution is a scalable architectural template: a semantic expert that preserves MLLM priors, a generation expert that learns sparse geometric flow matching, and a shared attention interface through which semantic and geometric tokens remain communicative. The gaps identified here---larger model scale, higher spatial resolution, finer evaluators, unified 3D-native encoders, and stronger dense visual conditioning---define a concrete path toward the next generation of 3D-native multimodal foundation models.

\section{Conclusion}

We presented \textbf{EVA01}, a unified framework that integrates 3D mesh understanding, generation, and multi-turn editing within a single Mixture-of-Transformers architecture.
By decoupling semantic understanding from geometric generation via a dual-expert design with shared global attention, EVA01 transfers multimodal priors from a pre-trained MLLM backbone to the 3D domain, bridging the semantic-geometric alignment gap under limited 3D supervision.

EVA01 achieves state-of-the-art native text-to-3D generation fidelity and enables context-aware, identity-preserving multi-turn 3D editing---a capability inaccessible to stateless reconstruction pipelines.
Our experiments yield three practical design principles for 3D-native MLLMs: grid-based sparse latents are necessary for geometric validity under unified sequence modeling; modality dropout and multi-stage curriculum training are essential for bridging textual and geometric manifolds; and high-fidelity finetuning sets the final fidelity ceiling after alignment is stabilized.

These findings establish a scalable architectural template for 3D-native multimodal models. The limitations discussed in Section~\ref{sec:limitations}---model scale, spatial resolution, evaluation fidelity, unified 3D-native encoders, and dense visual conditioning---define natural directions for future work.

\section*{Authors}

\noindent Zongyuan Yang, Mingjing Yi, Wanli Ma, Chenzhuo Fan, Bocheng Li, Baolin Liu, Yuke Lou, Yingde Song, Yongping Xiong, Zhengdong Guo, Shimu Wang.

\medskip
\noindent\textbf{Team Leaders.} Zhengdong Guo; Shimu Wang.
\quad
\textbf{Algorithm Leader.} Zongyuan Yang.

\noindent\textbf{Core Contributors.} Zongyuan Yang; Mingjing Yi; Wanli Ma.

\noindent\textbf{Contributors.} Chenzhuo Fan; Bocheng Li; Baolin Liu; Yuke Lou; Yingde Song; Yongping Xiong.

\clearpage
\bibliography{ref}

\end{document}